\documentclass{article}

\usepackage{ijcai25}
\usepackage{balance} 
\usepackage{graphicx} 
\usepackage{algorithm}
\usepackage{algorithmic}
\usepackage{latexsym}
\usepackage{amsmath}
\usepackage{amsthm}
\usepackage{float} 
\usepackage{booktabs}
\usepackage{enumitem}
\usepackage{color}
\usepackage{algorithm}
\usepackage{algorithmic}
\usepackage{bbm}
\usepackage{stmaryrd}
\usepackage{soul}
\usepackage{fancyhdr}
\usepackage{placeins}
\usepackage{url}
\usepackage{colortbl}
\usepackage[utf8]{inputenc}
\usepackage{makecell}
\usepackage{adjustbox}
\usepackage{tcolorbox}
\usepackage{mathtools}
\usepackage{multirow}
\usepackage{natbib}
\usepackage{multicol}
\usepackage{scontents}
\usepackage{lipsum}
\usepackage{tabularx}
\usepackage{array}
\newcolumntype{L}[1]{>{\raggedright\arraybackslash}p{#1}}

\theoremstyle{definition}

\usepackage{listings} 

\usepackage{amssymb}
\usepackage{soul}
\usepackage{array}
\usepackage{tikz}
\usetikzlibrary{decorations.pathreplacing}
\usetikzlibrary{calc}

\newcommand{\todo}[1]{{\textcolor{black}{#1}}} % Italic and blue color

\usepackage{times}
\usepackage[hidelinks]{hyperref}
\usepackage[utf8]{inputenc}
\usepackage[small]{caption}
\usepackage{graphicx}
\usepackage{amsfonts}
\usepackage{booktabs}
\usepackage[switch]{lineno}
\usepackage{xcolor}
\title{LogiDebrief: A Signal-Temporal Logic based Automated Debriefing Approach with Large Language Models Integration}

\author{
Zirong Chen$^1$
\and
Ziyan An$^1$
\and
Jennifer Reynolds$^2$
\and
Kristin Mullen$^2$
\and
Stephen Martini$^2$
\and
Meiyi Ma$^1$
\affiliations
$^1$Department of Computer Science, Vanderbilt University, Nashville, Tennessee 37235, USA \\
$^2$Metro Nashville Department of Emergency Communications, Nashville, Tennessee 37211, USA\\
\emails
\{zirong.chen, ziyan.an, meiyi.ma\}@vanderbilt.edu
\\
\{jennifer.reynolds, kristin.mullen, stephen.martini\}@nashville.gov
}

\begin{document}

\maketitle

\begin{abstract}

Emergency response services are critical to public safety, with 9-1-1 call-takers playing a key role in ensuring timely and effective emergency operations. To ensure call-taking performance consistency, quality assurance is implemented to evaluate and refine call-takers' skillsets. However, traditional human-led evaluations struggle with high call volumes, leading to low coverage and delayed assessments.  We introduce \textit{LogiDebrief}\todo{\footnote{More details: https://meiyima.github.io/angie.html}}, an AI-driven framework that automates traditional 9-1-1 call debriefing by integrating Signal-Temporal Logic (STL) with Large Language Models (LLMs) for fully-covered rigorous performance evaluation. LogiDebrief formalizes call-taking requirements as logical specifications, enabling systematic assessment of 9-1-1 calls against procedural guidelines. It employs a three-step verification process: (1) contextual understanding to identify responder types, incident classifications, and critical conditions; (2) STL-based runtime checking with LLM integration to ensure compliance; and (3) automated aggregation of results into quality assurance reports.  Beyond its technical contributions, LogiDebrief has demonstrated real-world impact. \todo{Successfully deployed at Metro Nashville Department of Emergency Communications}, it has assisted in debriefing 1,701 real-world calls, saving 311.85 hours of active engagement. Empirical evaluation with real-world data confirms its accuracy, while a case study and extensive user study highlight its effectiveness in enhancing call-taking performance.  

% Successfully deployed at local 9-1-1 call center\footnote{To maintain a double-blind review, we omit the city name. It is a mid-sized U.S. city (around 700k population) with a 9-1-1 call center employing around 80 staff.}, it has assisted in debriefing 1,244 real-world calls, saving 238.43 hours of active engagement. In a user study with real-world trainees, 92.59\% of participants preferred LogiDebrief over human-led debriefing.

\end{abstract}

\section{Introduction}

% Emergency response services are fundamental to public safety and are charged with protecting human lives, property, and the environment. At the core of these operations, 9-1-1 call-takers are the first point of contact in crises, where their decisions can directly impact response times, operational efficiency, and life-saving outcomes. Given the high stakes of their role, maintaining the highest possible standard of performance is not just a goal but a necessity. To ensure consistent and effective call handling, many emergency communication centers have implemented quality assurance programs, which involve structured frameworks designed to systematically evaluate call-taker performance, enforce adherence to standardized protocols, and provide actionable feedback for continuous improvement. These programs rely on call reviews, guidecard cross-referencing, and strict protocol verification to optimize call-taker efficiency and enhance the overall effectiveness of emergency response services.

Emergency response services are vital to public safety, with 9-1-1 call-takers as the first point of contact in crises, directly influencing response times and life-saving outcomes. Given their critical role, maintaining high performance is essential. To ensure consistency, emergency communication centers implement quality assurance programs that evaluate call-taker performance, enforce protocols, and provide actionable feedback. These programs use call reviews, guidecard cross-referencing, and protocol verification to enhance efficiency and emergency response effectiveness.

Despite their critical role, quality assurance programs across the U.S. face challenges in providing timely feedback due to high call volumes and limited resources~\citep{apco2020,IAED_NASNA_2023}. For example, during peak periods in 2024, the NYC Fire Department handled up to 6,500 emergency calls daily~\citep{ny_fdny_issue_brief}, straining quality assurance personnel. As urban populations grow and emergency call volumes rise, these challenges intensify~\citep{ma2019data}. Funding shortages and staffing constraints further hinder timely quality reviews~\citep{g12afonso2021planning}.  Timely feedback is essential for effective quality assurance~\citep{adarkwah2021power, ahea2016value}. Delays reduce relevance, making it harder for call-takers to recall key details, address performance gaps, and reinforce best practices. Without prompt debriefing, quality assurance programs risk becoming bottlenecks, delaying critical insights needed to improve call-taker training and emergency response. If unaddressed, these challenges may compromise 9-1-1 call centers’ ability to maintain high-performance standards, ultimately affecting response times and life-saving interventions.

Given these demands, an automated system is urgently needed for effective emergency call debriefing. While LLMs have advanced natural language processing~\citep{wei2023empirical, rouzegar2024enhancing}, their application in this domain presents significant challenges. Our preliminary trials identify three key \textbf{challenges}:  
(1) \textit{Step-by-step reasoning} is crucial for evaluating call-taker performance, as emergency calls require strict procedural adherence. LLMs must not only understand context but also apply structured reasoning. While In-Context Learning (ICL) techniques, such as Chain-of-Thought (CoT) prompting~\citep{wei2022chain}, improve reasoning, studies show that LLMs still struggle with complex, high-stakes decision-making~\citep{miao2023selfcheck, huang2023large, kambhampati2024can}. Even advanced automatic reasoning methods~\citep{zelikman2022star} exhibit weaknesses in scenarios requiring rigorous procedural verification~\citep{wu2024comparative, mccoy2024language}.  
(2) \textit{Cross-document retrieval and reference} remains another challenge. Call-takers rely on many complicated procedural documents, including guidecards, policies, and emergency protocols. While Retrieval-Augmented Generation (RAG)~\citep{lewis2020retrieval} assists by fetching external documents, its accuracy is inconsistent, often retrieving outdated or irrelevant information and struggling with multi-document synthesis~\citep{shi2023replug, shuster2022language}. Retrieval failures can result in missing critical protocols, significantly impacting evaluation.  
(3) The \textit{complex nature of emergency calls} further complicates debriefing, as a single call may span multiple protocols (e.g., a motor vehicle accident may begin under police protocols but escalate to medical and fire due to injuries or hazards). Some calls exceed 20 minutes, making evaluation even more challenging. Combining ICL and RAG often results in excessively long prompts that exceed optimal context windows. Empirical studies~\citep{weng2024mastering, dong2024exploring, an2024does, kuratov2024babilong} show that longer prompts degrade performance, leading to incomplete reasoning, ignored context, and lower factual consistency, as illustrated in Figure~\ref{fig:motivating}.

In this paper, we introduce \textit{LogiDebrief}, the first framework, to our knowledge, designed to automatically and effectively assist in 9-1-1 call-taking debriefing. LogiDebrief integrates logic-enhanced reasoning with LLMs' language understanding, providing an effective approach to evaluating call-taker performance.  Unlike traditional ICL methods that rely on lengthy prompts, LogiDebrief first collaborates with domain experts to decompose call-taking requirements into signal-temporal logic (STL) specifications~\citep{maler2004monitoring}. During runtime checking, LLMs function as independent evaluators within STL to verify compliance. Once verification is complete, LogiDebrief aggregates results, generates quality assurance forms, and delivers actionable feedback with tailored explanations. This automated, just-in-time debriefing process enhances call-taker training and improves emergency response effectiveness.

Our \textbf{technical innovations} and \textbf{contributions} are:  
(1) We introduce \textit{LogiDebrief}, a novel framework that automates 9-1-1 call-taking debriefing by integrating rigorous logic-based verification with LLM-powered analysis. (2) We decompose and formalize call-taking manuals into logic specifications through expert collaboration, ensuring standardized procedural verification, and improving consistency and reliability in 9-1-1 call debriefing.  
(3) We design an STL-integrated framework that seamlessly integrates LLMs as modular functions, enhancing procedural compliance while reducing the reliance on complex and lengthy prompts.  
(4) We empirically evaluate LogiDebrief’s performance through extensive experiments with real-world data, demonstrating its effectiveness in delivering accurate and reliable debriefing assessments.  
(5) We conduct a real-world case study and a user study (see user study in Appendix~\ref{app:user}) under practical deployment. The findings confirm that LogiDebrief is an effective tool for improving call-taking performance and training in emergency response settings.

% We validate our approach through a real-world case study, demonstrating LogiDebrief’s ability to detect procedural deviations and enhance call-taker training. A detailed user study further supports these findings (see Appendix).

% Beyond technical advancements, LogiDebrief also delivers the following \textbf{social impacts}: (1) LogiDebrief has been successfully deployed at the local 9-1-1 call center and integrated into the training session of both active call-takers and ongoing trainees. (2) To the date of this paper, LogiDebrief has assisted in debriefing 1,244 real-world calls and technically has saved over 238 working hours for the local call center. (3) LogiDebrief supports debriefing over 200 call cases across different responder departments, different call types, and different life-threatening situations. (4) In a user study, \todo{xxx} of participants would prefer LogiDebrief over human debriefing in terms of actionability, comprehensiveness, helpfulness, and self-explanation. (5) LogiDebrief has the potential to assist emergency communication centers across the US with limited staffing by automating the debriefing process.

Beyond its technical advancements, LogiDebrief delivers significant \textbf{social impact}:  
(1) Developed in collaboration with researchers and governmental agencies, LogiDebrief has been successfully deployed at \todo{Metro Nashville Department of Emergency Communications (MNDEC)}. It is now integrated into training programs for both active call-takers and trainees.  
(2) To date, it has assisted in debriefing 1,701 real-world calls, saving an estimated 311.85 working hours.  
(3) It facilitates debriefing for more than 200 call scenarios, covering various responder departments, call types, and life-threatening situations.  
(4) LogiDebrief has the potential to scale nationwide, offering automated debriefing solutions to emergency communication centers, particularly those operating in resource-constrained environments.

% (4) In a user study with 27 real-world trainees, 92.59\% participants preferred LogiDebrief over human debriefing in actionability, comprehensiveness, helpfulness, and self-explanation.

\section{Motivating Study}

% Through in-depth discussions with the local Department of Emergency Communication Center and a manual review of 1,244 past calls and their associated debriefing results, we identified critical limitations in the current practices of 9-1-1 call centers that demand urgent attention and underscore the urgent need for improved automated debriefing solutions capable of effectively handling the procedural intricacies of emergency call evaluations:

Through discussions with \todo{MNDEC} and a manual review of 1,244 past calls and debriefing results, we identified critical limitations in current 9-1-1 call center practices.

\noindent \textbf{Insufficient Call Review Coverage and Delayed Feedback} Emergency dispatch centers manage overwhelming call volumes daily, making comprehensive quality reviews increasingly difficult. At the local level, only 3.32\% of calls undergo manual review, covering approximately 2,000 to 2,300 calls per month. Nationwide, review rates often remain below 10\% due to resource constraints~\citep{apco2020,IAED_NASNA_2023}.  Debriefing a single call takes an average of 11.5 minutes, leading to backlogs that delay actionable insights and further strain quality assurance personnel.

\begin{figure}[h]
    \centering
    \includegraphics[width=\linewidth]{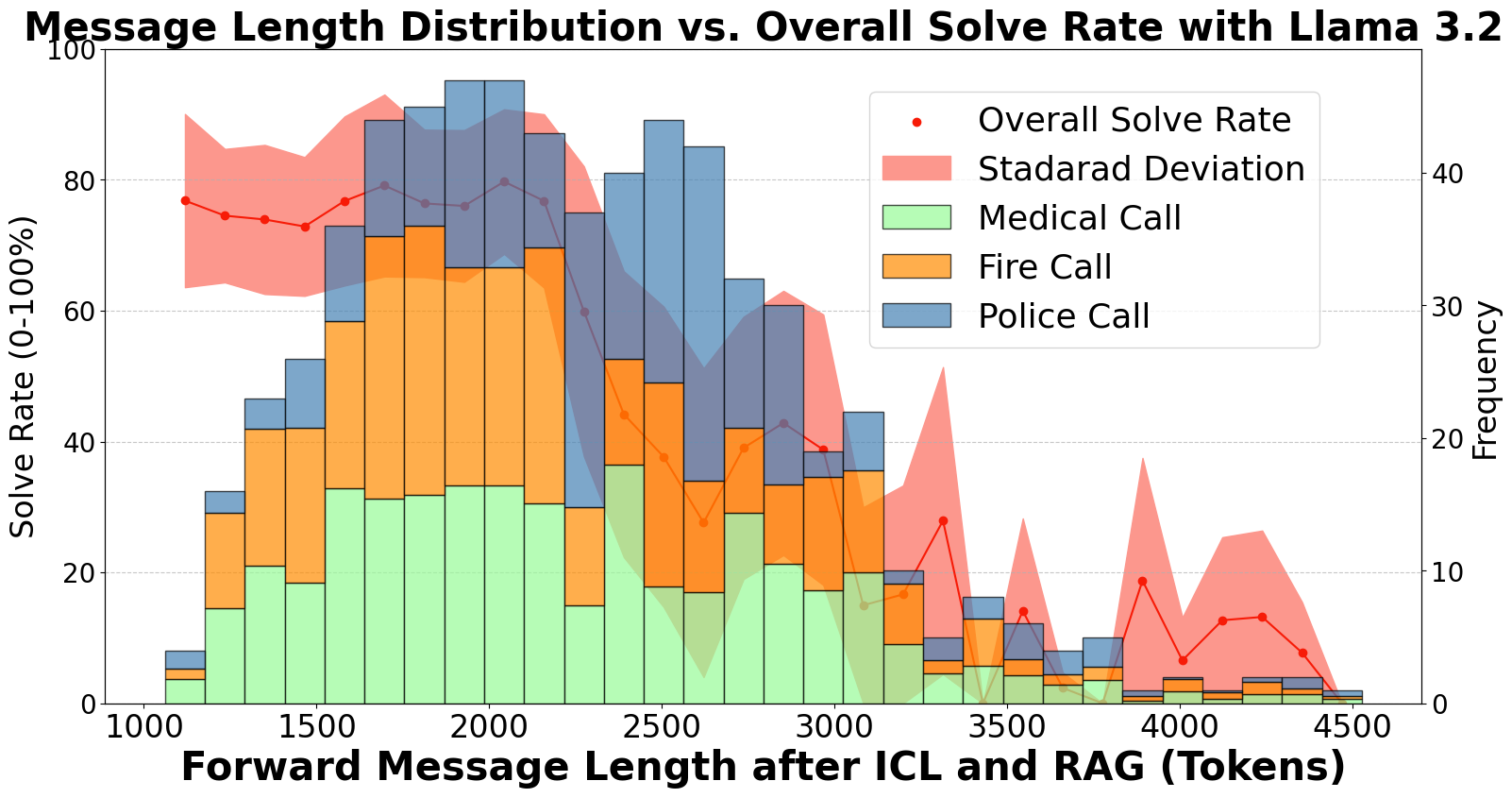}
    \caption{\small{Message Length vs. One-shot Solve Rate with Llama 3.2. This figure shows the relationship between prompt length (in tokens) and solve rate, segmented by primary call categories (Medical, Fire, and Police). Shaded regions indicate standard deviations.}}
    \label{fig:motivating}
\end{figure}

% \noindent \textbf{Challenges in LLM-Based Debriefing Workflows} We evaluated debriefing workflows using real-world samples with recent LLMs, including Llama 3.2~\citep{llama}, integrating ICL and RAG. Forwarded messages included step-by-step instructions, learning examples, and vectorized call-taking requirements. Our analysis compared LLM-generated debriefing results to ground truth while assessing performance relative to message length. Consistent with~\citep{weng2024mastering}, solve rates declined significantly for complex 9-1-1 calls requiring multiple guidecards, as shown in Figure~\ref{fig:motivating}. Call complexity increases with message length as extended prompts reference multiple protocols, impairing reasoning and lowering solve rates. Calls typically generate 1,800–3,000 tokens, where performance drops from 78\% to 42\%, with long-tail cases falling to 10\%. This highlights the challenge of handling complex conversations within LLMs' optimal context windows, limiting effective emergency call debriefing.

% Call complexity often correlates with message length: as the forwarded prompts (x-axis) extend, the call grows more complex, referencing multiple protocols. These excessively long prompts impair reasoning, lower solve rates, and hinder rigorous emergency call debriefing.

\noindent \textbf{Challenges in LLM-Based Debriefing Workflows} We evaluated debriefing workflows using real-world samples with recent LLMs, including Llama 3.2~\citep{llama}, integrating ICL and RAG. Forwarded messages included step-by-step instructions, learning examples, and vectorized call-taking requirements. Our analysis compared LLM-generated debriefing results to ground truth while assessing performance relative to message length.  As shown in Figure~\ref{fig:motivating}, calls typically generate 1,800 to 3,000 tokens, where performance drops approximately from 78\% to 42\%, with long-tail cases falling to 10\%. These results underscore the challenge of maintaining accuracy as call scenarios grow more intricate and involve broader procedural references and checks, as they are also consistent with~\citep{weng2024mastering, dong2024exploring, an2024does, kuratov2024babilong}.

\section{Problem Formulation and System Overview}

% This section first provides an overview of Deb9-1-1. Next, it introduces the preliminaries of the debriefing problems. Finally, it explores the technical aspects of Deb9-1-1 in more detail.

% \subsection{Overview}
% Call debriefing is a structured process for evaluating emergency call-handling performance. It involves systematically reviewing past calls to assess whether call-takers adhered to established protocols. This process ensures compliance with best practices, identifies potential areas for improvement, and enhances emergency response effectiveness. A comprehensive debriefing framework must account for the dynamic nature of conversations, filter relevant requirements based on call context, and provide an objective evaluation of procedural adherence. To facilitate automated call debriefing, LogiDebrief systematically analyzes past emergency calls by cross-referencing them with predefined call-taking criteria. The process begins by interpreting the call's context and filtering out irrelevant procedural requirements. Next, it monitors the call interactions against the applicable criteria to verify compliance. Finally, it aggregates the evaluation results and compiles them into a structured quality assurance form, providing clear and standardized feedback.

\begin{figure}[ht]
    \centering
    \includegraphics[width=\linewidth]{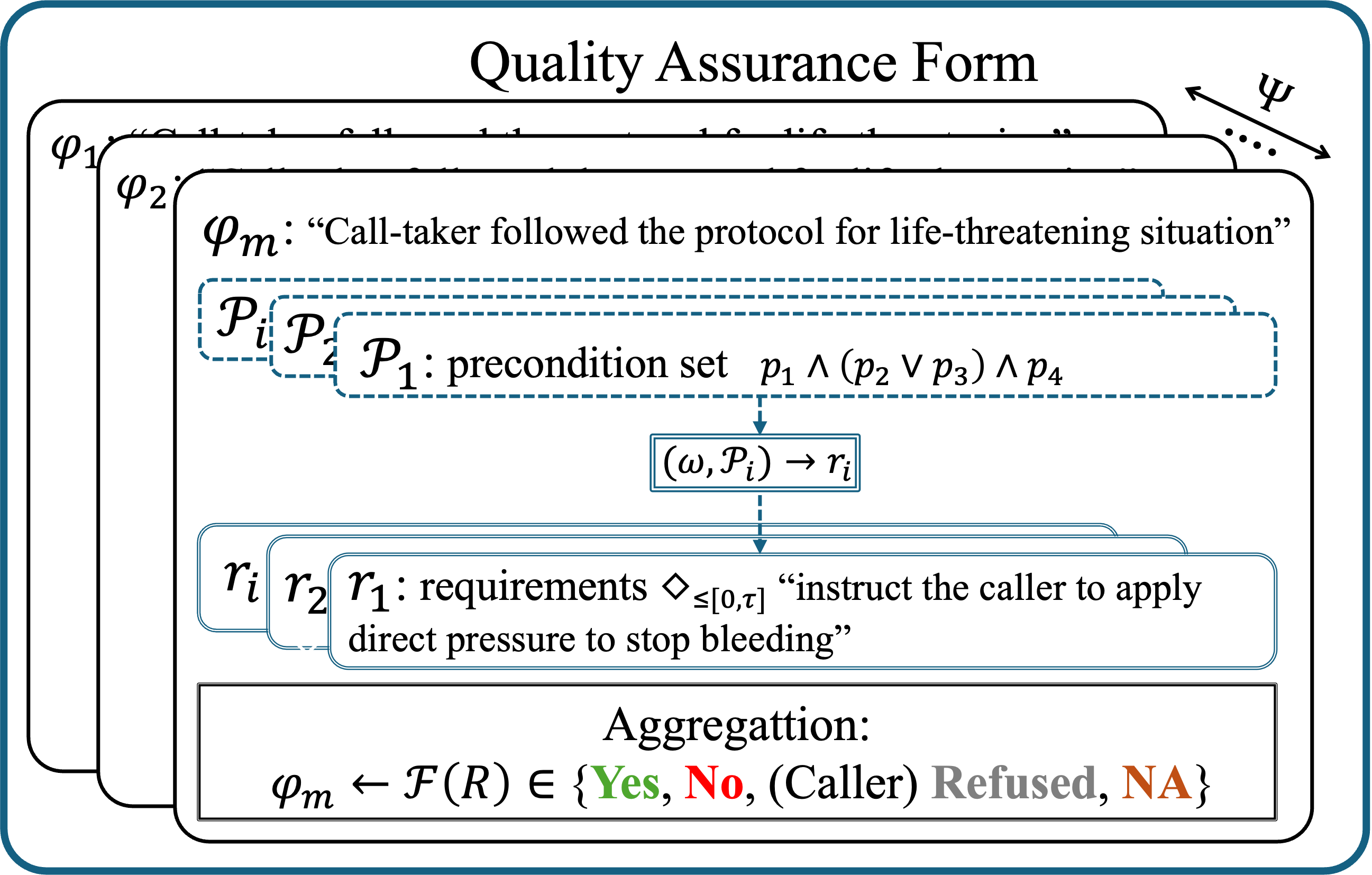}
    \caption{\small{Illustration of the Quality Assurance Form. Each form $\Psi$ consists of multiple checks ($\varphi$) derived from call-taking protocols. Requirements ($r$) are validated based on preconditions ($\mathcal{P}$) using extracted conversational signals ($\omega$). Aggregated results classify compliance such as Yes, No, (Caller) Refused, and NA.}}
    \label{fig:data}
\end{figure}

Call debriefing evaluates emergency call-handling performance by reviewing past calls for protocol adherence, ensuring compliance, identifying improvements, and enhancing emergency response. A robust framework must account for conversational dynamics, filter relevant requirements, and assess procedural adherence objectively. LogiDebrief automates this by interpreting call context, filtering irrelevant requirements, verifying compliance, and aggregating results into a quality assurance form with clear, standardized feedback. Following this, we formulate the 9-1-1 call debriefing problem, as more details are shown in Figure \ref{fig:data}. A \textbf{9-1-1 call} is a structured dialogue between call-taker $(a)$ and caller $(b)$, represented as: $\omega_{(ab)}\coloneqq \langle a_1, b_1, a_2, b_2, \dots, a_t, b_t \rangle$. Where $t$ is the number of the conversational turns, the call-taker's utterances and caller's utterances are defined correspondingly as, $\omega_{(a)} \coloneqq \langle a_1, \dots, a_t \rangle$ and $\omega_{(b)} \coloneqq \langle b_1, \dots, b_t \rangle$. 9-1-1 call-taking documents contain \textbf{requirements} $\mathcal{R}=\{r_1, r_2, \dots, r_p\}$. that call-takers must meet while handling an emergency call. Any \( r_i \in \{ \top, \bot\} \) is a predicate. $r_i$ is associated with a set of \textbf{preconditions} $\{\mathcal{P}_i \mid{r_i}\}$. Only when the entire \( \mathcal{P}_i \) holds given a conversational signal $\omega$, formally, $\mathbb{I}(\mathcal{P}_i \mid{r_i})$, will the corresponding \( r_i\) be applied for checks. The \textbf{quality assurance forms} $\Psi$ is based on the responder departments required for an emergency call, including fire, police, and medical. Each $\Psi$ consists of multiple checks $\Psi=\{\varphi_1, \varphi_2, \dots, \varphi_k\}.$ And $\forall \varphi \in \Psi, \varphi \in \{\text{Yes}, \text{No}, \text{(Caller) Refused}, \text{NA}\}$. Any $\varphi$ is aggregated from multiple associated requirements, formally written as $\varphi = \mathcal{F}(\mathcal{R})$, where $\mathcal{F}$ is the aggregation function, and $\mathcal{R}=\{r_1, r_2, \dots\}$ with each $r_i$ inferred from its precondition $\mathcal{P}_i$. 

% LogiDebrief assesses a call-taker's performance in past calls by cross-referencing relevant manuals and verifying adherence to established call-taking criteria. It processes past calls as temporal signals and retrieves STL-formalized requirements for runtime evaluation. During this phase, LogiDebrief utilizes LLMs' natural language understanding capabilities to enhance analysis. It first filters out irrelevant requirements and independently verifies compliance by scanning the extracted signals. Finally, it compiles the quality assurance form with aggregated results and structured template-based explanations.

% LogiDebrief assesses a call-taker's performance in past calls by cross-referencing relevant manuals and verifying adherence to established call-taking criteria. It processes past calls as temporal signals and retrieves STL-formalized requirements for runtime evaluation. During this phase, LogiDebrief first understands the call context with the assistance of LLMs and filters out irrelevant requirements. Then it monitors the call signal with STL specifications. Lastly, it aggregates the results and populates a quality assurance form with structured template-based explanations.

% utilizes LLMs' natural language understanding capabilities to enhance analysis. It first filters out irrelevant requirements and independently verifies compliance by scanning the extracted signals. Finally, it compiles the quality assurance form with aggregated results and structured template-based explanations. 

\section{Methodology}

\begin{figure*}[t]
    \centering
    \includegraphics[width=\linewidth]{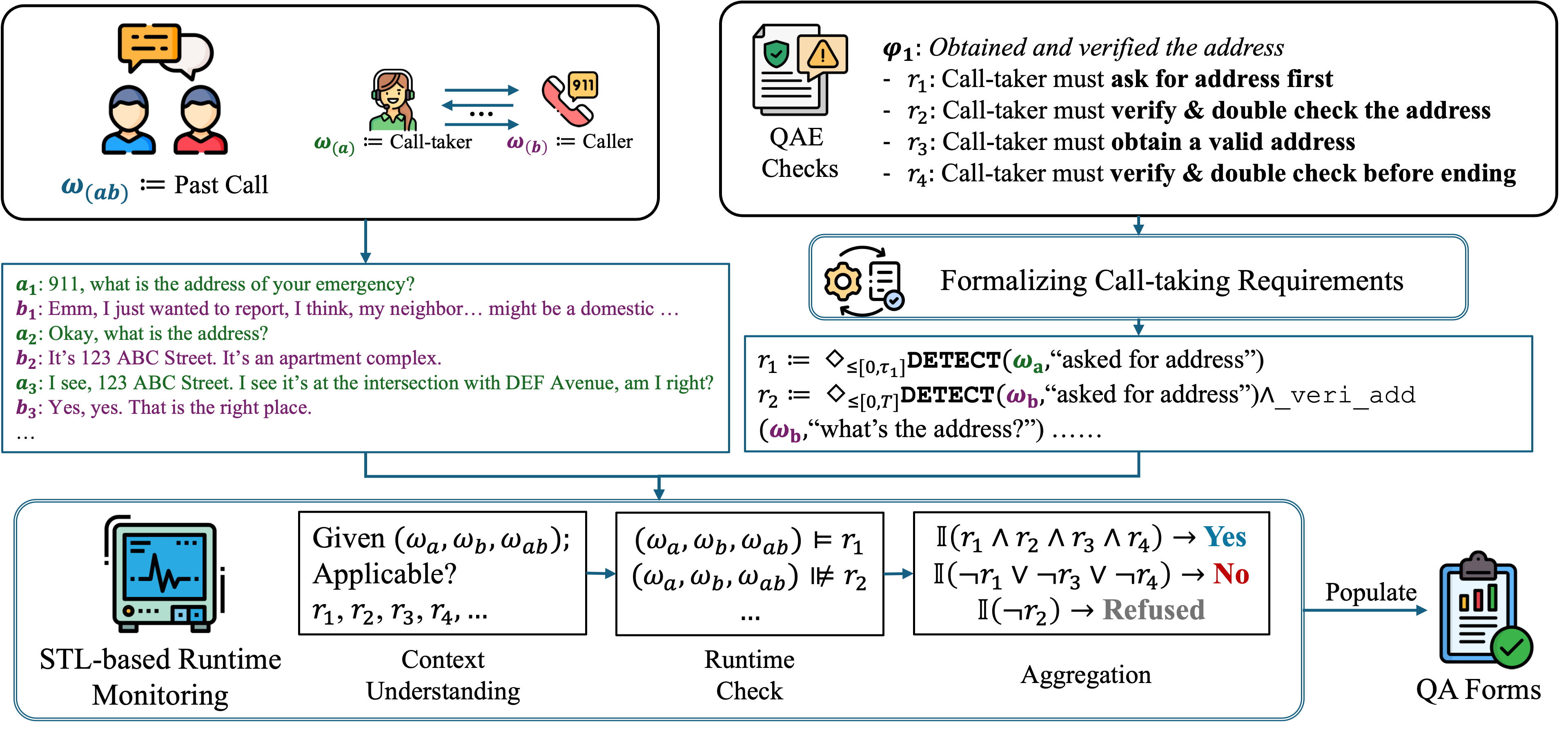}
    \caption{\small{Overview of the LogiDebrief Workflow. It evaluates call-taker performance by analyzing past calls against formalized requirements. It extracts conversational signals $\omega_{(a)}$ and $\omega_{(b)}$ from past calls, then applies quality assurance evaluation (QAE) checks. During runtime monitoring with LLMs, it finalizes applicable checks based on call context, checks the compliance of each check, and aggregates results into a quality assurance form with actionable template-based feedback.}}
    \label{fig:overview}
\end{figure*}

% In this section, we introduce LogiDebrief's workflow in more detail. LogiDebrief employs a structured approach to evaluate 9-1-1 call-taker performance by systematically analyzing call logs against established procedural guidelines. It integrates formalized call-taking requirements (Section \ref{sec:formalizing}) with runtime monitoring (Section \ref{sec:runtime}) to ensure compliance and identify potential gaps in adherence to emergency response protocols. LogiDebrief follows a three-step runtime checking process. First, it establishes context by determining the appropriate responders (Section \ref{sec:context}.1), identifying call types, assessing critical situations, and filtering out inapplicable checks to ensure relevant evaluations. Next, it conducts runtime verification using defined functions and structured logic-based rules to assess adherence to procedural guidelines (Section \ref{sec:check}.2). Finally, it aggregates results into a structured quality assurance report (Section \ref{sec:aggregate}.3), highlighting compliance, identifying deviations, and providing actionable feedback for call-taker training and performance improvement.

This section outlines LogiDebrief's workflow for 9-1-1 call debriefing. It systematically analyzes past calls against procedural guidelines, integrating formalized call-taking requirements (Section~\ref{sec:formalizing}) with runtime monitoring (Section~\ref{sec:runtime}) to ensure compliance and identify gaps. LogiDebrief follows a 3-step runtime checking process: (1) Establishing context by identifying responders, call types, and critical situations while filtering inapplicable checks (Section~\ref{sec:context}.1). (2) Conducting runtime verification through logic-based rules to assess procedural adherence (Section~\ref{sec:check}.2). (3) Aggregating results into a quality assurance form, highlighting compliance, deviations, and actionable feedback for training (Section~\ref{sec:aggregate}.3). Scalability and complexity considerations are further discussed in Appendix~\ref{app:complex}.

\subsection{Formalizing Call-taking Requirements}
\label{sec:formalizing}

% Through extensive discussions with the quality assurance team at the local 9-1-1 call center, we conducted a thorough review and reconstruction of existing call-taking requirements with domain expertise. By systematically analyzing each requirement, we identified and formalized their corresponding precondition sets, cataloging 2,215 distinct requirements, each with a detailed specification of its preconditions. These requirements span 57 general 9-1-1 call types, covering incidents such as heart problems, drowning, structure fires, and burglary. Additionally, we identified six critical protocols for life-threatening situations, including airway control, Automated External Defibrillator (AED) usage, bleeding control or tourniquet application, Cardiopulmonary Resuscitation (CPR), childbirth, and obstructed airways.

% These requirements are carefully decomposed and curated into Signal-Temporal Logic specifications along with their corresponding preconditions, with existing translation tools like~\citep{chen2022cityspec}. For example, under an animal bite call, if and only if the patient is bitten by snakes, the call-taker should alert the caller not to elevate the extremity. Otherwise, the caller should always elevate the extremity instead. This can be formally written as: $\Diamond_{[0, T]}$``call-taker should warn caller not to elevate the extremity'' with precondition $p_1 \wedge p_2$, where $p_1$ is ``the call is regarding an animal bite'' and $p_2$ is ``the patient is bitten by a snake''. 

Through discussions with the quality assurance team at \todo{MNDEC}, we systematically reviewed and reconstructed existing call-taking requirements with domain expertise. By analyzing each requirement, we identified and formalized 2,215 distinct requirements, each specifying its preconditions. These span 57 general 9-1-1 call types, including heart problems, drowning, structure fires, and burglary, as well as six critical life-threatening protocols: airway control, Automated External Defibrillator (AED) usage, bleeding control, Cardiopulmonary Resuscitation (CPR), childbirth, and obstructed airways.  Requirements were decomposed into STL specifications with preconditions using existing translation tools\todo{~\citep{cosler2023nl2spec, chen2022cityspec, chen2023cityspecs, chen2022intelligent}}. For instance, in an animal bite case, if and only if the patient was bitten by a snake, the call-taker should instruct the caller \textit{not} to elevate the extremity; otherwise, elevation is advised. This is formally expressed as \( \Diamond_{[0, T]} \) ``call-taker should warn caller not to elevate the extremity,'' with precondition \( p_1 \wedge p_2 \), where \( p_1 \) represents ``the caller reporting an animal bite,'' and \( p_2 \) denotes ``the patient was bitten by a snake.''

\subsection{STL-Based Runtime Monitoring with LLMs}
\label{sec:runtime}

% We integrate LLMs into runtime monitoring by encapsulating short, one-shot prompted LLM calls within functions defined in Signal Temporal Logic (STL), as examples showcased in Table \ref{tab:specs}. These functions serve over the conversational signal \( \omega \), enabling dynamic reasoning about call-taking compliance. Three major steps are involved in the overall debriefing process: (1) understanding the context, (2) runtime checking, and (3) aggregating the results. 

% We disentangle excessively lengthy procedural explanations and retrieval augmentation in prompts by leveraging STL and embedding independently prompted LLM calls as functions to efficiently handle modular tasks. This structured approach enhances rigor and interpretability while reducing prompt complexity. Runtime examples are in Table~\ref{tab:specs}. These functions operate over the conversational signal \( \omega \), enabling dynamic reasoning for assessing call-taking compliance. The debriefing process follows three key steps: (1) understanding the context, (2) runtime checking, and (3) aggregating results. Refer to Appendix for prompting details.

We disentangle lengthy procedural explanations and retrieval augmentations in prompts by embedding \textbf{\textit{independent modularized}} LLM calls as functions into STL. This approach extends STL's rigor, interpretability, and effectiveness~\citep{bae2019bounded,ergurtuna2022automated, an2025combining} while reducing prompt complexity. Runtime examples are in Table~\ref{tab:specs}. These functions operate over the conversational signal \( \omega \), enabling dynamic reasoning for assessing call-taking compliance. The debriefing process follows three key steps: (1) understanding the context, (2) runtime checking, and (3) aggregating results. Refer to Appendix \ref{app:prompt} for prompting \todo{and fail-safe} details.

% We integrate LLMs into runtime monitoring by embedding independently prompted LLM calls within Signal Temporal Logic (STL) functions to handle more agile and straightforward tasks. Runtime examples are shown in Table~\ref{tab:specs}. These functions operate over the conversational signal \( \omega \), enabling dynamic reasoning for call-taking compliance. The debriefing process consists of three key steps: (1) understanding the context, (2) runtime checking, and (3) aggregating results. Refer to Appendix for prompting details.

\subsubsection{Step 1: Understanding the Context}
\label{sec:context}

Understanding the context of a 9-1-1 call is crucial for procedural adherence. Each emergency requires a structured assessment, including identifying responder departments, classifying the incident type, and recognizing life-threatening situations requiring immediate intervention. These factors shape the call-taker's approach, determining the sequence of questions, instructions, and protocols. See Appendix~\ref{app:algo} for the algorithmic description.

% \noindent \textbf{\textit{Determining the Responders}}  The first step in contextualizing the call involves \textit{identifying the required responders}, denoted as \( \hat{\mathrm{R}} \). Since emergency services are categorized into three primary responders, fire, police, and medical, the system iterates through each category independently, analyzing the conversation for relevant indicators. We define the \texttt{SCENE} function. It returns \( \top \) if a relevant scene description is detected and \( \bot \) otherwise:  
% \begin{equation}
% \texttt{SCENE}(\omega, \text{responders}) \coloneqq 
% \Diamond_{[0,T]}  \underbrace{\big(\omega(t) \models \text{responders}\big)}_{\text{LLM Prompts}} 
% \end{equation}
% Our system iterates through each of the three scenes and obtains all applied ones:
% \begin{equation}
% \forall \mathrm{r} \in \{\mathrm{fire}, \mathrm{police}, \mathrm{medical}\}, \quad \texttt{SCENE}(\omega, \mathrm{r}) \rightarrow \hat{\mathrm{R}} \cup \{\mathrm{r}\}.
% \end{equation}
% Each responder category is evaluated separately, and the final set \( \hat{\mathrm{R}} \) consists of all departments for which the function evaluates to \( \top \), ensuring precise identification of the required emergency services.  

\noindent \textbf{\textit{Determining the Responders}} The first step in contextualizing a call is \textit{identifying the required responders}, denoted as \( \hat{\mathrm{R}} \). Emergency services, fire, police, and medical, are analyzed independently for relevant indicators. We define the \texttt{SCENE} function, which returns \( \top \) if a relevant scene description is detected and \( \bot \) otherwise:
\begin{equation}
\texttt{SCENE}(\omega, \text{responders}) \coloneqq 
\Diamond_{[0,T]}  \underbrace{\big(\omega(t) \models \text{responders}\big)}_{\text{LLM Prompts}} 
\end{equation}
Iterating through the three categories, the system obtains all applicable responders:
\begin{equation}
\forall \mathrm{r} \in \{\mathrm{fire}, \mathrm{police}, \mathrm{medical}\}, \quad \texttt{SCENE}(\omega, \mathrm{r}) \rightarrow \hat{\mathrm{R}} \cup \{\mathrm{r}\}
\end{equation}
The final set \( \hat{\mathrm{R}} \) includes all departments where the function evaluates to \( \top \), ensuring accurate identification of required emergency services.

\noindent \textbf{\textit{Identifying the Call Types}} After determining the required responders, the next step is to \textit{classify the call type}, such as structure fires or heart problems, denoted as \( \hat{\mathrm{T}} \). Instead of evaluating all incident types indiscriminately, classification is conditioned on the identified responders. We define the \texttt{TYPE} function, which classifies the call by analyzing its semantic context. It returns \( \top \) if sufficient evidence supports the specified call type and \( \bot \) otherwise:
\begin{equation}
\texttt{TYPE}(\omega, \text{type}) \coloneqq 
\Diamond_{[0,T]}  \underbrace{\big(\omega(t) \models \text{type}\big)}_{\text{LLM Prompts}}  
\end{equation}
The system iterates over call types relevant to the confirmed responders:
\begin{equation}
\forall \mathrm{t} \in \mathrm{Call \ Types} \mid{\hat{\mathrm{R}}}, \quad \texttt{TYPE}(\omega, \mathrm{t}) \rightarrow \hat{\mathrm{T}} \cup \{\mathrm{t}\}.
\end{equation}
Thus, only incident types relevant to the identified responder departments are considered. E.g., if only police response is required, fire- or medical-related incidents such as structure fires or diabetic emergencies are excluded. The final result includes all incident types where the function evaluates to \( \top \), ensuring accurate classification.

\noindent \textbf{\textit{Alerting Critical Situations}} Following up, LogiDebrief identifies predefined \textit{life-threatening situations} requiring immediate intervention. Each of the 6 critical conditions, airway control, AED, bleeding control, CPR, childbirth, and obstructed airways, is analyzed independently. We define the \texttt{CRITICAL} function to check if any of these conditions apply to a given call \( \omega \):
\begin{equation}
\texttt{CRITICAL}(\omega, \text{flag}) \coloneqq 
\Diamond_{[0,T]}  \underbrace{\big(\omega(t) \models \text{flag}  \big)}_{\text{LLM Prompts}}
\end{equation}
This process is formally expressed as:
\begin{equation}
\forall \mathrm{c} \in \mathrm{Criticals}_{\times6}, \quad \texttt{CRITICAL}(\omega, \mathrm{c}) \rightarrow \hat{\mathrm{C}} \cup \{\mathrm{c}\}
\end{equation}
where \( \hat{\mathrm{C}} \) represents the set of flagged critical situations. LogiDebrief iterates through all six conditions, ensuring that any applicable life-threatening scenario is detected and appropriate emergency protocols are triggered without delay.

\begin{table*}[t]
    \small
    \centering
\resizebox{\textwidth}{!}{%
\begin{tabular}{{|m{9.5cm} p{13.25cm}|}}
\hline
\multicolumn{2}{|l|}{\textbf{$\varphi_{\text{address}}$: Address Check}}      \\ \hline \hline
\multicolumn{1}{|l|}{$r_1$: \normalsize{call-taker asked for address in first $\tau_1$ turns.}}         &   \normalsize{$\Diamond_{\leq [0, \tau_1]} \texttt{DETECT}(\omega_a, \text{`ask address'})$}      \\ \hline
\multicolumn{1}{|l|}{$r_2$: \normalsize{caller provided a valid address.}}         &     
\makecell[l]{
\normalsize{$\Diamond_{[0,T]} \big( \texttt{DETECT}(\omega_b, \text{`provide address'}) \wedge$} \\
\normalsize{$\texttt{veri\_add}(\texttt{answer}(\omega_b, \text{`what's the address?'})) \big)$}
} \\ \hline
\multicolumn{1}{|l|}{$r_3$: \normalsize{call-taker verified the obtained address with nearby geo-info.}}         &     
\makecell[l]{
\normalsize{$\Box_{[0,T]} \big( \texttt{DETECT}(\omega_b, \text{`provide address'}) \wedge \texttt{veri\_add}(\texttt{answer}(\omega_b, \text{`what's the address?'})) \rightarrow$} \\
\normalsize{$\Diamond_{\leq \tau} \big( \texttt{DETECT}(\omega_a, \text{`double checks address'}) \wedge$} \\
\normalsize{$\texttt{veri\_add}(\texttt{answer}(\omega_a, \text{`what's address?'}), \texttt{answer}(\omega_b, \text{`what's address?'}) ) \big) \big)$}
}
\\ \hline 
\multicolumn{1}{|l|}{$r_4$: \normalsize{call-taker verified the address again before getting disconnected.}}         &    
\makecell[l]{
\normalsize{$\Diamond_{[T -\tau_2, T]} \big( \texttt{DETECT}(\omega_a, \text{`double checks address'}) \wedge$} \\
\normalsize{$\texttt{veri\_add}(\texttt{answer}(\omega_a, \text{`what's address?'}))$}
}     \\ \hline \hline

\multicolumn{2}{|l|}{\textbf{$\varphi_{\text{name}}$: Caller Name Check} \& \textbf{$\varphi_{\text{phone}}$: Caller Phone Check}}         \\ \hline \hline
\multicolumn{1}{|l|}{$r_1$: \normalsize{call-taker asked for both first and last name / phone number.}}         &    
\makecell[l]{
\normalsize{$\Diamond_{[0,T]} \big( \texttt{DETECT}(\omega_a, \text{`ask for full name / phone number'})$}
}     \\ \hline
\multicolumn{1}{|l|}{$r_2$: \normalsize{caller provided both first and last name / phone number.}}         &   
\makecell[l]{
\normalsize{$\Diamond_{[0,T]} \big( \texttt{DETECT}(\omega_b, \text{`provide full name / phone number'})$}
}      \\ \hline
\multicolumn{1}{|l|}{$r_3$: \normalsize{call-taker followed up with caller's name / phone number.}}         &     
\makecell[l]{
\normalsize{$\Box_{[0,T]} \big( \texttt{DETECT}(\omega_b, \text{`provides name / phone'}) \rightarrow$} \\
\normalsize{$\Diamond_{\leq \tau_1} \texttt{DETECT}(\omega_a, \text{`follows up on name / phone'}) \big)$}
}    \\ \hline \hline
\multicolumn{2}{|l|}{$\varphi_i \in \Psi$: \textbf{Conditional Checks}} \\ \hline
\multicolumn{1}{|l|}{
\makecell[l]{
$r_1$:\normalsize{ if the scene is potentially not safe for police officers,} \\
\normalsize{call-taker obtained scene safety info.}
}
}         &       
\makecell[l]{
\normalsize{$\Diamond_{\leq [0,\tau_3]} \texttt{DETECT}(\omega_{ab}, \text{`scene safety info obtained'})$}
}
\\ \hline
\multicolumn{1}{|l|}{
\makecell[l]{
$r_2$:\normalsize{ if the patient is an infant and not breathing,} \\
\normalsize{call-taker should do [\textit{{infant CPR}}: \textit{step 1, step 2, ...}].}
}
}         &    
\makecell[l]{
\normalsize{$\Diamond_{\leq [0,\tau_4]} \texttt{DETECT}(\omega_a, \text{`call-taker instructs [\textit{{infant CPR}}: \textit{step 1, step 2, ...}]'})$}
}
     \\ \hline
\multicolumn{1}{|l|}{
\makecell[l]{
$r_3$:\normalsize{ if the call involves medical emergency,} \\
\normalsize{call-taker should check if patient is breathing normally.}
}
}         &      
\makecell[l]{
\normalsize{$\Diamond_{\leq [0,\tau_5]} \texttt{DETECT}(\omega_a, \text{`checked patient breathing'})$}
}
\\ \hline
\multicolumn{1}{|l|}{
\makecell[l]{
$r_4$:\normalsize{ if there is any suspicious vehicle spotted on the scene,}\\ 
\normalsize{call-taker should ask for detailed vehicle descriptions.}
}
}      &     
\makecell[l]{
\normalsize{$\Diamond_{\leq [0,\tau_6]} \texttt{DETECT}(\omega_a, \text{`asked for vehicle description'})$}
}
\\ \hline
\multicolumn{1}{|l|}{
\makecell[l]{
$r_5$: \normalsize{if the roadway hazard is blocking the traffic, call-taker} \\ 
\normalsize{should warn caller not to move the hazard by themselves.}
}
}         &    
\makecell[l]{
\normalsize($\Diamond_{\leq [0,\tau_7]} \texttt{DETECT}(\omega_a, \text{`warn caller not to move the hazard'})$)
}\\ \hline
\multicolumn{1}{|l|}{
\makecell[l]{
$r_6$: \normalsize{if the caller reports an odor, call-taker should warn caller} \\
\normalsize{to avoid using energized equipment that could cause a spark.}
}}         &    
\makecell[l]{
\normalsize{$\Diamond_{\leq [0,\tau_8]} \texttt{DETECT}(\omega_a, \text{`warn caller not to use energized equipment'})$}
}
\\ \hline
\end{tabular}}
    \caption{\normalsize{A runtime example of both conditional and unconditional checks in natural languages and STL specifications. Each of the conditional checks $\varphi$ satisfies Equation \ref{eq:precondition}. All $\tau$ are adaptable hyper-parameters for different call-taking requirements.}}
    \label{tab:specs}
\end{table*}

\noindent \textbf{\textit{Finalizing Checks}} After determining \( \hat{\mathrm{R}} \), \( \hat{\mathrm{T}} \), and \( \hat{\mathrm{C}} \), LogiDebrief finalizes checks. These checks dynamically adjust based on scene information denoted as $\Gamma(.)$; e.g., medical-related forms verify patient assessment, while police-related forms ensure scene safety.
\begin{equation}
    \Psi = \Gamma(\hat{\mathrm{R}}), \quad \Psi = \{\varphi_1, \varphi_2, \dots, \varphi_k\}
\end{equation}
While scene information defines the form’s structure, refinements based on \( \hat{\mathrm{T}} \) and \( \hat{\mathrm{C}} \) update specific requirements without introducing new structural checks. Each check \( \varphi \) in \( \Psi \) is linked to a set of requirements \( \{r_1, r_2, \dots\} \), which adapt based on the emergency scenario:
\begin{equation}
    \forall \varphi \in \Psi, \quad \varphi \leftarrow \varphi \oplus \Delta(\hat{\mathrm{T}}, \hat{\mathrm{C}})
    \label{eq:dynamic}
\end{equation}
where \( \oplus \) updates requirements using relevant call-taking manuals while preserving form structure. \( \Delta(\hat{\mathrm{T}}, \hat{\mathrm{C}}) \) applies context-specific refinements; e.g., a cardiac arrest call updates breathing assessments to explicitly verify chest compressions. After applying \( \oplus \Delta(\hat{\mathrm{T}}, \hat{\mathrm{C}}) \) to each \( \varphi \in \Psi \), the final quality assurance form \( \Psi \) is obtained.

% After having $\Psi$, before we check the runtime, we iterate through each $\varphi \in \Psi$ to check the compliance. For each \( \varphi \), we retrieve the associated requirement set \( \{r_1, r_2, \dots, r_i\} \) and its corresponding precondition set \( \{\mathcal{P}_1, \mathcal{P}_2, \dots, \mathcal{P}_i\} \). To assist this process, we define \texttt{SCAN} function to scan $\omega$ with a precondition to verify if the precondition holds in $\omega$. It returns \( \top \) if the precondition holds at any time \( t \) within the observation window, and \( \bot \) otherwise. This can be formally written as:
% \begin{equation}
% \texttt{SCAN}(\omega, \text{precondition}) \coloneqq \Diamond_{[0,T]} \underbrace{ \big(\omega(t) \models \text{precondition} \big)}_{\text{LLM Prompts}}
% \end{equation}
% \begin{equation}
% \texttt{SCAN}(\omega, p), \quad \forall p \in \{\mathcal{P} \mid{r}, \ \forall r \in \varphi\}.
% \end{equation}
% The indication of the entire $\mathcal{P}_i$ is represented as:
% \begin{equation}
% \mathbb{I}(\mathcal{P} \mid{r}) = \bigwedge_{p \in \{\mathcal{P}\mid{r}\}} \texttt{SCAN}(\omega, p).
% \end{equation}
% Based on the precondition check, we determine whether the requirement \( \mathcal{R}_i \) should be monitored. If the preconditions do not hold, \( \mathcal{R}_i \) is skipped and will not participate in runtime monitoring. Ultimately, a quality assurance form $\Psi$ should satisfy:
% \begin{equation}
%     \forall \varphi \in \Psi, \forall r \in \varphi \quad \mathbb{I}(\mathcal{P} \mid{r}) = \top
%     \label{eq:precondition}
% \end{equation}

After obtaining \( \Psi \), we iterate through each \( \varphi \in \Psi \) to check compliance. For each \( \varphi \), we retrieve its associated requirements \( \{r_1, r_2, \dots, r_i\} \) and corresponding preconditions \( \{\mathcal{P}_1, \mathcal{P}_2, \dots, \mathcal{P}_i\} \). To facilitate this process, we define the \texttt{SCAN} function, which verifies whether a precondition holds in \( \omega \). It returns \( \top \) if the condition is met at any time \( t \) within the observation window, and \( \bot \) otherwise:
\begin{equation}
\texttt{SCAN}(\omega, \text{precondition}) \coloneqq \Diamond_{[0,T]} \underbrace{ \big(\omega(t) \models \text{precondition} \big)}_{\text{LLM Prompts}}
\end{equation}
The overall precondition evaluation is represented as:
\begin{equation}
\forall p \in \{\mathcal{P} \mid{r}, \ \forall r \in \varphi\}, \quad \texttt{SCAN}(\omega, p)
\end{equation}
\begin{equation}
\mathbb{I}(\mathcal{P} \mid{r}) = \bigwedge_{p \in \{\mathcal{P}\mid{r}\}} \texttt{SCAN}(\omega, p)
\end{equation}
If the preconditions do not hold, requirement \( \mathcal{R}_i \) is skipped and excluded from runtime monitoring. Ultimately, a quality assurance form \( \Psi \) should satisfy:
\begin{equation}
    \forall \varphi \in \Psi, \forall r \in \varphi \quad \mathbb{I}(\mathcal{P} \mid{r}) = \top
    \label{eq:precondition}
\end{equation}

% \begin{equation}
% \mathbb{I}(\mathcal{P}_i \mid{r_i}) \rightarrow r_i.
% \end{equation}
% \begin{equation}
% \neg\mathbb{I}(\mathcal{P}_i) \rightarrow r_i = \mathrm{NA}.
% \end{equation}

% \paragraph{Detecting an Action}  
% This function verifies whether a specific \textit{action} occurs within the conversation signal \( \omega \). It returns \( \top \) if the action is present at any time \( t \) within the observation window, and \( \bot \) otherwise.  
% \begin{equation}
% \texttt{DETECT}(\omega, \text{action}) \coloneqq \Diamond_{[0,T]} \underbrace{ \big(\omega(t) \models \text{action} \big)}_{\text{LLM Prompts}}
% \end{equation}

% \paragraph{Scanning a Precondition}
% This function scans the $\omega$ with a precondition to verify if the precondition holds in $\omega$. It returns \( \top \) if the precondition holds at any time \( t \) within the observation window, and \( \bot \) otherwise.
% \begin{equation}
% \texttt{SCAN}(\omega, \text{precondition}) \coloneqq \Diamond_{[0,T]} \underbrace{ \big(\omega(t) \models \text{precondition} \big)}_{\text{LLM Prompts}}
% \end{equation}

% \todo{we first get $\mathcal{R}_i$ then for each $r \in \mathcal{R}_i$ has a $\mathcal{P}$, then we do following...}

\subsubsection{Step 2: Checking the Runtime}
\label{sec:check}

We define functions to verify requirements. The \texttt{DETECT} function checks whether a specific \textit{action} occurs within the conversation signal \( \omega \), returning \( \top \) if detected at any time \( t \) within the observation window and \( \bot \) otherwise:
\begin{equation}
\texttt{DETECT}(\omega, \text{action}) \coloneqq \Diamond_{[0,T]} \underbrace{ \big(\omega(t) \models \text{action} \big)}_{\text{LLM Prompts}}
\end{equation}
Additional non-STL functions further analyze \( \omega \):
\begin{equation}
    \texttt{answer}(\omega, \text{query}) \rightarrow a.
\end{equation}
The \texttt{answer} function integrates LLMs for question-answering, retrieving the most relevant response \( a \) from \( \omega \). If no answer is found, it returns an empty string.

\begin{equation}
    \texttt{veri\_add}(\text{addresses}) \rightarrow \{\top, \bot\}
\end{equation}
The \texttt{veri\_add} function utilizes Google Geocoding and Places API to validate addresses. If given an empty string, it returns \( \top \) by default. For a \textit{single address}, it returns \( \top \) if successfully located on a map; otherwise, \( \bot \). For \textit{two addresses}, it returns \( \top \) if geographically close; otherwise, \( \bot \). These functions are embedded in STL specifications for just-in-time runtime verification after each call. Examples of their integration are shown under each \( \varphi \) in Table~\ref{tab:specs}.

\subsubsection{Step 3: Aggregating the Results}
\label{sec:aggregate}

After having satisfaction for each $r_i \in \mathcal{R}$, we aggregate the result to populate:
\begin{equation}
    \varphi \leftarrow \mathcal{F}(\mathcal{R}), \quad \varphi \in \{\text{Yes}, \text{No}, \text{Refused}, \text{NA}\}
\end{equation}
This aggregation step is instructed back to trainees with template-based natural language generation as explanations for the populated results, e.g., ``Your overall evaluation at this check is NO, because you missed $r$.''

\noindent \textbf{Unconditional Checks} According to the call-taking manual, three checks, \(\varphi_1, \varphi_2, \varphi_3\) in Table~\ref{tab:specs}, are always examined regardless of call context: \textit{address}, caller \textit{name}, and phone \textit{number}, each with distinct aggregation rules.

The address check categorizes verification performance into three outcomes: (1) \textit{Yes}: The call-taker successfully requests (\( r_1 \)), verifies (\( r_3 \)), and reconfirms (\( r_4 \)) the address, and the caller provides a valid one (\( r_2 \)). For example, if a geocodable location is confirmed before call termination, the outcome is \textit{Yes}. (2) \textit{No}: The call-taker fails to ask for (\( r_1 \)), verify (\( r_3 \)), or reconfirm (\( r_4 \)) the address, regardless of caller response. If geographical verification is neglected, the call is classified as \textit{No}. (3) \textit{Refused}: If the caller explicitly refuses to provide an address (\( r_2 \)), the outcome is \textit{Refused}, as the failure is beyond the call-taker’s control, e.g., a distressed caller declining to disclose their location. Formally:
\begin{equation}
 \varphi_{\text{address}} = 
\begin{cases}
    \text{Yes}, & \text{if } r_1 \wedge r_2 \wedge r_3 \wedge r_4 \\
    \text{No}, & \text{if } \neg r_1 \vee \neg r_3 \vee \neg r_4\\
    \text{Refused}, & \text{if } \neg r_2
\end{cases}
\end{equation}

The outcomes of caller name and phone checks classify the call-taker’s performance in collecting caller identity into three categories: (1) \textbf{Yes}: The call-taker correctly requests (\( r_1 \)), receives (\( r_2 \)), and follows up on (\( r_3 \)) the caller’s name and phone number, ensuring full compliance. For example, if all details are requested, provided, and confirmed, the outcome is \textit{Yes}. (2) \textbf{No}: The call-taker fails to request (\( r_1 \)) or follow up (\( r_3 \)) on the information, regardless of whether the caller provides it. If verification is omitted, the call is classified as \textit{No}. (3) \textbf{Refused}: The caller explicitly refuses to provide their name or phone number (\( r_2 \)), making the failure beyond the call-taker’s control, e.g., a caller declining to disclose their identity despite multiple requests. Formally:
\begin{equation}
\varphi_{\text{name}}, \varphi_{\text{phone}} =
\begin{cases}
    \text{Yes}, & \text{if } r_1\wedge r_2 \wedge r_3 \\
    \text{No}, & \text{if } \neg r_1\vee \neg r_3 \\
    \text{Refused}, & \text{if } \neg r_2
\end{cases}
\end{equation}

\noindent \textbf{Conditional Checks} The outcome of any conditional checks \( \varphi_i \) depends on the satisfaction of all monitored requirements \( r_i \):  
(1) \textbf{Yes}: All applicable conditional checks are met, meaning every monitored \( r_i \) returns \( \top \). For example, if the scene was unsafe (\( r_1 \)) and the call-taker obtained scene safety information, the outcome is \textit{Yes}.  
(2) \textbf{No}: At least one monitored requirement fails (\( r_i = \bot \)). For instance, if a medical emergency is detected (\( r_3 \)), but the call-taker fails to check the patient's breathing, the outcome is \textit{No}.  
(3) \textbf{NA}: No requirements were monitored due to unsatisfied preconditions.
\begin{equation}
\varphi_i =
\begin{cases}
    \text{Yes}, & \text{if } \forall r \in \varphi_i, r = \top \\
    \text{No}, & \text{if } \exists r \in \varphi_i, r = \bot \\
    \text{NA}, & \text{if } \varphi_i = \emptyset
\end{cases}
\end{equation}

\section{Evaluation}

% both quantitatively and qualitatively 

LogiDebrief is a pioneering AI-driven system designed to automate 9-1-1 call debriefing. Given its novelty, limited literature exists to guide its evaluation. To ensure a comprehensive assessment, we evaluate its effectiveness through both quantitative benchmarking and real-world case studies. In addition, we conducted a user study, detailed in Appendix \ref{app:user}, to further assess its impact on enhancing call-taking performance.

\begin{table*}[ht]
\centering
\small
\resizebox{\textwidth}{!}{%
\renewcommand{\arraystretch}{1.25}
\begin{tabular}{|ccc|ccc|ccccccccc|}
\hline
\multicolumn{3}{|c|}{\multirow{4}{*}{}}                                                                                                                      & \multicolumn{3}{c|}{\multirow{2}{*}{\textsc{Real-World}}}                                                                                      & \multicolumn{9}{c|}{\textsc{Emulation}}                                                                                                                                                                                                                                                                                                                                                                                                                                                                              \\ \cline{7-15} 
\multicolumn{3}{|c|}{}                                                                                                                                       & \multicolumn{3}{c|}{}                                                                                                                                       & \multicolumn{3}{c|}{$\alpha \gets$25}                                                                                                                                            & \multicolumn{3}{c|}{$\alpha \gets$50}                                                                                                                                            & \multicolumn{3}{c|}{$\alpha \gets$75}                                                                                                                       \\ \cline{4-15} 
\multicolumn{3}{|c|}{}                                                                                                                                       & \multicolumn{2}{c|}{$\forall \varphi \in \Psi$}                                                                                   & \multirow{2}{*}{$\Psi$} & \multicolumn{2}{c|}{$\forall \varphi \in \Psi$}                                                                                   & \multicolumn{1}{c|}{\multirow{2}{*}{$\Psi$}} & \multicolumn{2}{c|}{$\forall \varphi \in \Psi$}                                                                                   & \multicolumn{1}{c|}{\multirow{2}{*}{$\Psi$}} & \multicolumn{2}{c|}{$\forall \varphi \in \Psi$}                                                                                   & \multirow{2}{*}{$\Psi$} \\ \cline{4-5} \cline{7-8} \cline{10-11} \cline{13-14}
\multicolumn{3}{|c|}{}                                                                                                                                       & \multicolumn{1}{c|}{\textit{\textit{\small{Unconditional}}}} & \multicolumn{1}{c|}{\textit{\textit{\small{Conditional}}}} &                         & \multicolumn{1}{c|}{\textit{\textit{\small{Unconditional}}}} & \multicolumn{1}{c|}{\textit{\textit{\small{Conditional}}}} & \multicolumn{1}{c|}{}                        & \multicolumn{1}{c|}{\textit{\textit{\small{Unconditional}}}} & \multicolumn{1}{c|}{\textit{\textit{\small{Conditional}}}} & \multicolumn{1}{c|}{}                        & \multicolumn{1}{c|}{\textit{\textit{\small{Unconditional}}}} & \multicolumn{1}{c|}{\textit{\textit{\small{Conditional}}}} &                         \\ \hline \hline
\multicolumn{1}{|c|}{\multirow{12}{*}{\begin{tabular}[c]{@{}c@{}}\rotatebox{90}{Typical LLMs}\end{tabular}}} & \multicolumn{1}{c|}{\multirow{3}{*}{Llama 3.2}}   & Vanilla & \multicolumn{1}{c|}{37.11$\pm$16.93}                             & \multicolumn{1}{c|}{36.24$\pm$16.10}                           & 7.12$\pm$5.03           & \multicolumn{1}{c|}{29.51$\pm$8.01}                              & \multicolumn{1}{c|}{31.69$\pm$4.36}                            & \multicolumn{1}{c|}{8.97$\pm$7.57}           & \multicolumn{1}{c|}{25.12$\pm$9.23}                              & \multicolumn{1}{c|}{32.23$\pm$10.72}                           & \multicolumn{1}{c|}{7.90$\pm$6.47}           & \multicolumn{1}{c|}{35.62$\pm$17.17}                             & \multicolumn{1}{c|}{32.50$\pm$9.82}                            & 8.85$\pm$7.03           \\ \cline{3-15} 
\multicolumn{1}{|c|}{}                                                                         & \multicolumn{1}{c|}{}                             & RAG     & \multicolumn{1}{c|}{40.99$\pm$13.72}                             & \multicolumn{1}{c|}{36.04$\pm$17.09}                           & 10.85$\pm$7.17          & \multicolumn{1}{c|}{34.04$\pm$9.49}                              & \multicolumn{1}{c|}{31.51$\pm$8.37}                            & \multicolumn{1}{c|}{8.66$\pm$3.27}           & \multicolumn{1}{c|}{30.29$\pm$12.39}                             & \multicolumn{1}{c|}{28.22$\pm$9.43}                            & \multicolumn{1}{c|}{9.56$\pm$5.79}           & \multicolumn{1}{c|}{34.19$\pm$13.27}                             & \multicolumn{1}{c|}{29.88$\pm$10.61}                           & 7.71$\pm$6.70           \\ \cline{3-15} 
\multicolumn{1}{|c|}{}                                                                         & \multicolumn{1}{c|}{}                             & ICL+RAG & \multicolumn{1}{c|}{75.39$\pm$14.33}                             & \multicolumn{1}{c|}{78.24$\pm$13.33}                           & 26.85$\pm$15.89         & \multicolumn{1}{c|}{71.73$\pm$8.94}                              & \multicolumn{1}{c|}{74.12$\pm$7.53}                            & \multicolumn{1}{c|}{20.60$\pm$14.14}         & \multicolumn{1}{c|}{81.17$\pm$9.15}                              & \multicolumn{1}{c|}{77.61$\pm$11.11}                           & \multicolumn{1}{c|}{22.06$\pm$14.14}         & \multicolumn{1}{c|}{73.13$\pm$18.29}                             & \multicolumn{1}{c|}{78.09$\pm$11.64}                           & 24.11$\pm$12.87         \\ \cline{2-15} 
\multicolumn{1}{|c|}{}                                                                         & \multicolumn{1}{c|}{\multirow{3}{*}{Gemma 2}}     & Vanilla & \multicolumn{1}{c|}{37.81$\pm$17.15}                             & \multicolumn{1}{c|}{27.56$\pm$12.45}                           & 11.62$\pm$7.49          & \multicolumn{1}{c|}{36.94$\pm$9.20}                              & \multicolumn{1}{c|}{31.10$\pm$11.05}                           & \multicolumn{1}{c|}{9.44$\pm$3.00}           & \multicolumn{1}{c|}{34.89$\pm$13.94}                             & \multicolumn{1}{c|}{39.38$\pm$9.76}                            & \multicolumn{1}{c|}{11.08$\pm$10.82}         & \multicolumn{1}{c|}{41.43$\pm$15.11}                             & \multicolumn{1}{c|}{45.61$\pm$11.78}                           & 11.18$\pm$6.55          \\ \cline{3-15} 
\multicolumn{1}{|c|}{}                                                                         & \multicolumn{1}{c|}{}                             & RAG     & \multicolumn{1}{c|}{38.16$\pm$15.90}                             & \multicolumn{1}{c|}{29.52$\pm$11.06}                           & 11.50$\pm$10.64         & \multicolumn{1}{c|}{39.93$\pm$11.75}                             & \multicolumn{1}{c|}{33.38$\pm$12.73}                           & \multicolumn{1}{c|}{8.73$\pm$7.73}           & \multicolumn{1}{c|}{32.37$\pm$11.66}                             & \multicolumn{1}{c|}{34.14$\pm$15.15}                           & \multicolumn{1}{c|}{14.19$\pm$8.32}          & \multicolumn{1}{c|}{42.89$\pm$17.86}                             & \multicolumn{1}{c|}{44.75$\pm$12.58}                           & 10.53$\pm$7.22          \\ \cline{3-15} 
\multicolumn{1}{|c|}{}                                                                         & \multicolumn{1}{c|}{}                             & ICL+RAG & \multicolumn{1}{c|}{71.88$\pm$15.94}                             & \multicolumn{1}{c|}{73.91$\pm$14.55}                           & 30.97$\pm$11.22         & \multicolumn{1}{c|}{72.05$\pm$9.95}                              & \multicolumn{1}{c|}{73.91$\pm$13.81}                           & \multicolumn{1}{c|}{25.18$\pm$14.14}         & \multicolumn{1}{c|}{72.61$\pm$15.03}                             & \multicolumn{1}{c|}{73.69$\pm$10.12}                           & \multicolumn{1}{c|}{25.81$\pm$11.58}         & \multicolumn{1}{c|}{69.82$\pm$10.23}                             & \multicolumn{1}{c|}{68.43$\pm$12.64}                           & 23.18$\pm$9.57          \\ \cline{2-15} 
\multicolumn{1}{|c|}{}                                                                         & \multicolumn{1}{c|}{\multirow{3}{*}{DeepSeek-v3}} & Vanilla & \multicolumn{1}{c|}{45.22$\pm$15.29}                             & \multicolumn{1}{c|}{52.47$\pm$14.51}                           & 11.57$\pm$6.44          & \multicolumn{1}{c|}{48.32$\pm$19.53}                             & \multicolumn{1}{c|}{50.41$\pm$13.64}                           & \multicolumn{1}{c|}{12.74$\pm$11.59}         & \multicolumn{1}{c|}{43.71$\pm$13.09}                             & \multicolumn{1}{c|}{46.53$\pm$11.98}                           & \multicolumn{1}{c|}{11.54$\pm$7.74}          & \multicolumn{1}{c|}{48.53$\pm$13.77}                             & \multicolumn{1}{c|}{45.00$\pm$14.59}                           & 13.13$\pm$9.45          \\ \cline{3-15} 
\multicolumn{1}{|c|}{}                                                                         & \multicolumn{1}{c|}{}                             & RAG     & \multicolumn{1}{c|}{59.75$\pm$11.39}                             & \multicolumn{1}{c|}{57.49$\pm$18.66}                           & 19.11$\pm$8.34          & \multicolumn{1}{c|}{53.64$\pm$16.84}                             & \multicolumn{1}{c|}{58.24$\pm$17.61}                           & \multicolumn{1}{c|}{12.68$\pm$8.89}          & \multicolumn{1}{c|}{56.29$\pm$11.46}                             & \multicolumn{1}{c|}{57.97$\pm$18.12}                           & \multicolumn{1}{c|}{13.40$\pm$7.38}          & \multicolumn{1}{c|}{57.74$\pm$13.76}                             & \multicolumn{1}{c|}{56.27$\pm$12.11}                           & 15.97$\pm$8.96          \\ \cline{3-15} 
\multicolumn{1}{|c|}{}                                                                         & \multicolumn{1}{c|}{}                             & ICL+RAG & \multicolumn{1}{c|}{86.21$\pm$12.95}                             & \multicolumn{1}{c|}{87.29$\pm$11.93}                           & 55.21$\pm$2.31          & \multicolumn{1}{c|}{86.55$\pm$13.33}                             & \multicolumn{1}{c|}{87.95$\pm$10.05}                           & \multicolumn{1}{c|}{52.67$\pm$13.47}         & \multicolumn{1}{c|}{84.49$\pm$13.95}                             & \multicolumn{1}{c|}{85.82$\pm$12.87}                           & \multicolumn{1}{c|}{54.54$\pm$9.43}          & \multicolumn{1}{c|}{86.19$\pm$11.71}                             & \multicolumn{1}{c|}{84.99$\pm$14.39}                           & 55.08$\pm$7.49          \\ \cline{2-15} 
\multicolumn{1}{|c|}{}                                                                         & \multicolumn{1}{c|}{\multirow{3}{*}{GPT-4o}}      & Vanilla & \multicolumn{1}{c|}{41.60$\pm$15.26}                             & \multicolumn{1}{c|}{52.59$\pm$13.72}                           & 10.54$\pm$8.87          & \multicolumn{1}{c|}{54.50$\pm$17.05}                             & \multicolumn{1}{c|}{55.72$\pm$11.90}                           & \multicolumn{1}{c|}{12.00$\pm$8.20}          & \multicolumn{1}{c|}{54.24$\pm$10.36}                             & \multicolumn{1}{c|}{56.43$\pm$11.17}                           & \multicolumn{1}{c|}{13.93$\pm$9.45}          & \multicolumn{1}{c|}{55.40$\pm$16.62}                             & \multicolumn{1}{c|}{59.41$\pm$11.14}                           & 12.95$\pm$4.25          \\ \cline{3-15} 
\multicolumn{1}{|c|}{}                                                                         & \multicolumn{1}{c|}{}                             & RAG     & \multicolumn{1}{c|}{54.51$\pm$18.10}                             & \multicolumn{1}{c|}{58.16$\pm$17.81}                           & 17.23$\pm$7.03          & \multicolumn{1}{c|}{56.67$\pm$13.98}                             & \multicolumn{1}{c|}{61.83$\pm$14.50}                           & \multicolumn{1}{c|}{21.54$\pm$10.52}         & \multicolumn{1}{c|}{62.84$\pm$10.49}                             & \multicolumn{1}{c|}{59.52$\pm$12.19}                           & \multicolumn{1}{c|}{15.49$\pm$6.72}          & \multicolumn{1}{c|}{56.57$\pm$15.54}                             & \multicolumn{1}{c|}{57.15$\pm$12.14}                           & 16.93$\pm$6.81          \\ \cline{3-15} 
\multicolumn{1}{|c|}{}                                                                         & \multicolumn{1}{c|}{}                             & ICL+RAG & \multicolumn{1}{c|}{88.16$\pm$10.65}                             & \multicolumn{1}{c|}{87.78$\pm$10.79}                           & 58.91$\pm$13.14         & \multicolumn{1}{c|}{88.31$\pm$11.58}                             & \multicolumn{1}{c|}{87.03$\pm$11.79}                           & \multicolumn{1}{c|}{57.13$\pm$13.98}         & \multicolumn{1}{c|}{85.02$\pm$13.71}                             & \multicolumn{1}{c|}{85.79$\pm$12.57}                           & \multicolumn{1}{c|}{54.13$\pm$14.46}         & \multicolumn{1}{c|}{87.11$\pm$12.04}                             & \multicolumn{1}{c|}{84.60$\pm$14.76}                           & 53.16$\pm$13.84         \\ \hline \hline
\multicolumn{1}{|c|}{\multirow{2}{*}{Reasoners}}                                              & \multicolumn{2}{c|}{Deepseek-r1}                            & \multicolumn{1}{c|}{86.84$\pm$10.18}                             & \multicolumn{1}{c|}{87.75$\pm$7.58}                            & 60.50$\pm$12.44         & \multicolumn{1}{c|}{88.88$\pm$10.62}                             & \multicolumn{1}{c|}{87.76$\pm$11.83}                           & \multicolumn{1}{c|}{60.20$\pm$14.48}         & \multicolumn{1}{c|}{86.57$\pm$13.19}                             & \multicolumn{1}{c|}{86.23$\pm$11.67}                           & \multicolumn{1}{c|}{58.91$\pm$12.10}         & \multicolumn{1}{c|}{86.09$\pm$11.33}                             & \multicolumn{1}{c|}{85.24$\pm$12.44}                           & 55.22$\pm$14.80         \\ \cline{2-15} 
\multicolumn{1}{|c|}{}                                                                         & \multicolumn{2}{c|}{OpenAI-o1}                              & \multicolumn{1}{c|}{88.21$\pm$9.79}                              & \multicolumn{1}{c|}{89.83$\pm$8.17}                            & 63.74$\pm$10.11         & \multicolumn{1}{c|}{87.63$\pm$12.37}                             & \multicolumn{1}{c|}{88.75$\pm$11.25}                           & \multicolumn{1}{c|}{59.62$\pm$15.13}         & \multicolumn{1}{c|}{87.50$\pm$12.50}                             & \multicolumn{1}{c|}{89.33$\pm$10.22}                           & \multicolumn{1}{c|}{60.93$\pm$17.83}         & \multicolumn{1}{c|}{88.45$\pm$11.55}                             & \multicolumn{1}{c|}{87.63$\pm$10.37}                           & 58.52$\pm$16.11         \\ \hline \hline
\multicolumn{1}{|c|}{\multirow{4}{*}{\textbf{LogiDebrief}}}                                    & \multicolumn{2}{c|}{Llama 3.2}                              & \multicolumn{1}{c|}{81.30$\pm$4.02}                              & \multicolumn{1}{c|}{88.62$\pm$7.63}                            & 58.80$\pm$6.40          & \multicolumn{1}{c|}{87.90$\pm$3.13}                              & \multicolumn{1}{c|}{88.78$\pm$6.33}                            & \multicolumn{1}{c|}{54.33$\pm$9.05}          & \multicolumn{1}{c|}{85.52$\pm$5.02}                              & \multicolumn{1}{c|}{87.22$\pm$7.11}                            & \multicolumn{1}{c|}{50.12$\pm$6.46}          & \multicolumn{1}{c|}{84.51$\pm$3.21}                              & \multicolumn{1}{c|}{87.96$\pm$6.33}                            & 51.52$\pm$5.76          \\ \cline{2-15} 
\multicolumn{1}{|c|}{}                                                                         & \multicolumn{2}{c|}{Gemma 2}                                & \multicolumn{1}{c|}{77.75$\pm$4.16}                              & \multicolumn{1}{c|}{80.78$\pm$6.12}                            & 59.24$\pm$2.35          & \multicolumn{1}{c|}{78.29$\pm$4.88}                              & \multicolumn{1}{c|}{79.70$\pm$4.49}                            & \multicolumn{1}{c|}{52.40$\pm$6.08}          & \multicolumn{1}{c|}{80.45$\pm$4.10}                              & \multicolumn{1}{c|}{81.11$\pm$5.00}                            & \multicolumn{1}{c|}{51.49$\pm$2.71}          & \multicolumn{1}{c|}{78.15$\pm$3.61}                              & \multicolumn{1}{c|}{80.98$\pm$3.52}                            & 52.24$\pm$8.14          \\ \cline{2-15} 
\multicolumn{1}{|c|}{}                                                                         & \multicolumn{2}{c|}{DeepSeek-v3}                            & \multicolumn{1}{c|}{92.63$\pm$5.08}                              & \multicolumn{1}{c|}{90.89$\pm$4.05}                            & 84.75$\pm$5.25          & \multicolumn{1}{c|}{93.90$\pm$4.10}                              & \multicolumn{1}{c|}{91.43$\pm$5.94}                            & \multicolumn{1}{c|}{86.36$\pm$6.14}          & \multicolumn{1}{c|}{92.68$\pm$6.51}                              & \multicolumn{1}{c|}{90.86$\pm$5.33}                            & \multicolumn{1}{c|}{82.25$\pm$6.35}          & \multicolumn{1}{c|}{91.10$\pm$4.38}                              & \multicolumn{1}{c|}{91.70$\pm$2.47}                            & 83.06$\pm$5.51          \\ \cline{2-15} 
\multicolumn{1}{|c|}{}                                                                         & \multicolumn{2}{c|}{GPT-4o}                                 & \multicolumn{1}{c|}{\textbf{95.93$\pm$4.07}}                     & \multicolumn{1}{c|}{\textbf{94.40$\pm$5.60}}                   & \textbf{94.33$\pm$5.67} & \multicolumn{1}{c|}{\textbf{94.39$\pm$5.61}}                     & \multicolumn{1}{c|}{\textbf{95.07$\pm$4.93}}                   & \multicolumn{1}{c|}{\textbf{94.84$\pm$5.16}} & \multicolumn{1}{c|}{\textbf{95.45$\pm$4.55}}                     & \multicolumn{1}{c|}{\textbf{95.38$\pm$4.62}}                   & \multicolumn{1}{c|}{\textbf{94.62$\pm$5.38}} & \multicolumn{1}{c|}{\textbf{94.49$\pm$5.51}}                     & \multicolumn{1}{c|}{\textbf{95.61$\pm$4.39}}                   & \textbf{94.04$\pm$5.96} \\ \hline
\end{tabular}
}
\caption{\small{Evaluation of LogiDebrief with \textsc{real-world} and \textsc{emulation} data compared with baselines. $\alpha$ is call-taker `proficiency levels': $\alpha$ percentage of the required actions are taken during the scripted emulation. At the $\varphi$ level, performance is evaluated per check $\varphi$. At the $\Psi$ level, a response is counted as correct only if the entire set is populated accurately. Performance is reported in multi-fold F-1 scores as $\%$.
}}
\label{tab:eval}
\end{table*}

Quantitatively, we investigate \textit{\textbf{how effectively LogiDebrief debriefs 9-1-1 calls}}. We first evaluate LogiDebrief on 1,244 real-world calls with debriefing results provided by quality assurance experts at \todo{MNDEC}. However, since professional 9-1-1 operators handle these calls, errors are rare, potentially leading to an inflated false positive rate. Additionally, this dataset lacks coverage of rare but critical incidents (e.g., aircraft crashes, nuclear leaks).  To address these limitations, we \textit{construct a diverse dataset} encompassing various call types and call-taker proficiency levels:  
(1) Defining all requirements with their preconditions (e.g., snake vs. non-snake bites);  
(2) \todo{Using LLMs to generate simulated 9-1-1 reports under role-play~\citep{chen2025sim911}}; and  
(3) \todo{Interacting with controlled actions~\citep{chen2024auto311}} where call-takers access only a percentage (\(\alpha\)) of requirements. Those scripted actions also generate the ground truth. 
 Here, \(\alpha\) represents familiarity level; higher \(\alpha\) values indicate greater adherence to required actions. For instance, in an animal bite emergency, a scripted call-taker may fail to ask about the animal type if \textit{``What type of animal caused the bite?''} is masked. We set \(\alpha = 25, 50, 75\) to simulate varying proficiency levels, totaling 13,200 calls with corresponding quality assurance forms. 
Performance is reported with F-1 scores for \(\{\text{Yes}, \text{No}, \text{Refused}, \text{NA}\}\) after multi-fold validation. Proprietary LLMs (GPT-4o, DeepSeek-v3-671B) and reasoning models (OpenAI-o1) are tested via API, while open-source and smaller LLMs run with 128 GB RAM, AMD Ryzen Threadripper Pro 7975WX, and NVIDIA RTX 6000 Ada.

Qualitatively, we focus on \textit{\textbf{how effectively LogiDebrief enhances call-taking performance in real-world settings}}. 
To assess this, we conduct a case study at \todo{MNDEC}. Additionally, we conducted a user study to further validate LogiDebrief’s effectiveness in enhancing call-taker training. Full details \todo{and complete results} are presented in Appendix \ref{app:user}.

% In this case, a call-taker was managing a rapidly evolving emergency that required coordination between multiple responder units. Due to the complexity of the situation, certain procedural confirmations were momentarily deferred to prioritize immediate response actions. While the existing quality assurance process provided a comprehensive review of the call, LogiDebrief identified these procedural deferrals and offered structured feedback to help reinforce best practices for handling similar cases in the future. 

\subsection{Effectiveness in Call Debriefing}

We evaluate LogiDebrief's performance following baseline setups:  (1) \textit{Vanilla LLMs}, where the full quality assurance form \( \Psi \) (Eq.~\ref{eq:dynamic}) is provided as input, and responses are generated directly.  (2) \textit{LLMs with RAG}, utilizing vectorized call-taking manuals as knowledge bases without logical structuring.  (3) \textit{LLMs with RAG+ICL}, combining RAG with Chain-of-Thought reasoning and Few-Shot examples for procedural explanations.  (4) \textit{Reasoning frameworks} tested with necessary step-by-step instructions.  We evaluate these setups using available LLM backends (Llama3.2-3B~\citep{llama}, Gemma2-9B~\citep{gemma2}, DeepSeek-v3-671B~\citep{deepseek}, and GPT-4o~\citep{openai2024gpt4o}) and Reasoners (OpenAI-o1-2024-12-17~\citep{openai2024o1} and DeepSeek-r1~\citep{deepseek2025r1}). See the Appendix \ref{app:prompt} for detailed prompting methods.

Tab.~\ref{tab:eval} presents key insights across real-world and emulation scenarios:  
(1) \textit{Vanilla LLMs underperform}, with F1 scores below 40\% (e.g., Llama 3.2 Vanilla: 37.11$\pm$16.93 in unconditional checks). This confirms that call debriefing requires multi-step validation and logical reasoning, which standard LLMs struggle with. Even RAG, which incorporates call-taking manuals, provides minimal improvement (e.g., Llama 3.2 RAG: 40.99$\pm$13.72), as manuals contain static rules without reasoning mechanisms, limiting generalization beyond simple lookups.    
(2) \textit{ICL+RAG improves step-by-step reasoning} and serves as the strongest LLM-based alternative for procedural verification. Thus, we consider it an approximation of LogiDebrief without STL. However, lacking explicit logical constraints, it remains prone to reasoning errors, particularly in strict procedural checks (e.g., GPT-4o ICL+RAG: 87.78$\pm$10.79).  
(3) \textit{Reasoning models provide only marginal improvements}, with DeepSeek-r1 and OpenAI-o1 achieving 88.21$\pm$9.79 and 86.84$\pm$10.18, respectively. While they exhibit better problem comprehension, they lack procedural enforcement. LogiDebrief surpasses all baselines by integrating STL, ensuring rigorous protocol adherence beyond what LLMs alone can achieve.  
\textit{Overall, LogiDebrief outperforms all baselines across real-world and emulation datasets, demonstrating that integrating formalized logic with LLM reasoning yields the most effective call debriefing performance.}

% (3) \textit{Reasoning models provide limited improvements}, with DeepSeek-r1 and OpenAI-o1 scoring 88.21$\pm$9.79 and 86.84$\pm$10.18, respectively. These models exhibit better problem comprehension but still fall short of LogiDebrief’s accuracy, particularly in complex decision validation.  The key advantage of LogiDebrief lies in its integration of STL, enabling structured, runtime-verified debriefing. 

% \textit{Across both real-world and emulation datasets, LogiDebrief consistently outperforms all baselines, demonstrating that integrating formal logic with LLM reasoning achieves superior call debriefing performance, even surpassing specialized reasoning models.}

\subsection{Case Study: LogiDebrief in the Field}  

We conducted a case study of LogiDebrief under its 4-week active engagement at \todo{MNDEC} (Nov 2024 – Jan 2025) under daily use and 2 training sessions. In the study, LogiDebrief cross-reviewed 1,244 calls alongside human debriefing and independently analyzed 457 calls. A total of 29 participants contributed, including 16 trainees, 5 active call-takers, and 8 training/quality assurance officers, providing 37 feedback entries.  We share the following findings:  (1) \textbf{\textit{Timeliness.}} Traditional quality assurance feedback is provided at the end of a shift, making it difficult for call-takers, who handle over 80 calls daily, to recall specific interactions. This delay reduces evaluation effectiveness and limits immediate skill reinforcement. LogiDebrief delivers just-in-time feedback, generating quality assurance reports in under 6 seconds per minute of call audio. Compared to the 11.5-minute manual review process, it reduces evaluation time to 4.45\% ($<$30 seconds) per call while maintaining accuracy, saving over estimated 311 working hours. A quality assurance officer noted:  \textit{``The feedback was quick and spot-on. It even caught the mistakes I made on purpose. This can really save a lot of time.''}  (2) \textbf{\textit{Higher Coverage.}} LogiDebrief boosted call review coverage by 73.96\% to 85.05\%, processing 1,701 more calls compared to previous 2,000 to 2,300 per 4 weeks under human efforts. (3) \textbf{\textit{Comprehensiveness.}} Traditional quality assurance often emphasizes errors without reinforcing correct practices. Feedback can be generic, making it harder for call-takers to extract actionable insights. LogiDebrief provides balanced assessments, highlighting both strengths and areas for improvement. Its STL-enhanced check offers step-by-step guidance, clarifying why specific actions were correct or required adjustment. One call-taker shared:  \textit{``It walked me through step by step instead of just flagging mistakes, so I knew exactly what went wrong and how to fix it.''}  In summary, LogiDebrief enhances call-taking performance by providing timely, accurate, and actionable feedback. By reducing evaluation time, increasing review coverage, and improving instructional clarity, it supports continuous learning and strengthens procedural consistency in emergency response.

\section{Related Work}

\textbf{Automated debriefing} is well-studied in education and medical training, where structured feedback enhances skill development. Intelligent tutoring systems provide adaptive feedback for language learning, STEM education, and problem-solving but focus on static assessments rather than real-time procedural evaluation~\citep{graesser2012intelligent, VanLehn2006}. In medical training, AI-assisted tools assess procedural adherence in surgical simulations and emergency medicine~\citep{mcgaghie2010critical, toews2021clinical}. However, emergency call-taking remains largely overlooked despite its need for timely feedback. \textbf{Large Language Models for procedural checks} face reliability challenges. Chain-of-Thought prompting \citep{wei2022chain} improves reasoning but does not ensure strict adherence, leading to hallucinations and missing steps~\citep{turpin2024language, ling2024deductive}. Retrieval-Augmented Generation (RAG) \citep{lewis2020retrieval} improves factual accuracy but cannot guarantee retrieving relevant procedural guidelines, making it unreliable for high-stakes verification~\citep{chen2024benchmarking, wang2024searching}. Self-verification improves consistency but lacks procedural rigor~\citep{zelikman2022star, chung2024scaling}, while longer prompts degrade multi-step procedural integration~\citep{weng2024mastering}. These limitations underscore the need for more robust verification. Extended related work is in Appendix \ref{app:related}.

\section{Summary}
% In this paper, we introduce LogiDebrief, the first AI-driven framework designed to automate and assist the 9-1-1 call-taking debriefing. LogiDebrief enables structured and scalable evaluation of call-taker performance by integrating logic-driven procedural verification with LLM-powered analysis.  Evaluation and user study results demonstrate that LogiDebrief effectively debriefs real-world 9-1-1 calls and further helps call-taking performance enhancement. 

% This work can help emergency communication centers with limited staffing by automating quality assurance and reducing manual review burdens. With over 6,000 emergency communication centers across the U.S., LogiDebrief presents an opportunity to improve call-taker training at scale. Beyond emergency response, the framework can be extended to structured compliance audits in fields such as medical triage, law enforcement, and legal training.

In this paper, we introduce LogiDebrief, the first AI-driven framework for automating and assisting 9-1-1 call-taking debriefing. Integrating logic-driven procedural verification with LLM-powered analysis, LogiDebrief enables rigorous call-taker performance evaluation. Evaluation and case studies confirm its effectiveness in debriefing real-world 9-1-1 calls and enhancing call-taking performance.  

This work can support emergency communication centers with limited resources by assisting with quality assurance and reducing manual debriefing burdens. With over 6,000 emergency communication centers across the US, it offers an effective approach for call-taker performance enhancement. Beyond emergency response, LogiDebrief's framework can potentially extend to structured compliance audits in other training spaces, such as medical triage and law enforcement.  

\bibliographystyle{named}
\bibliography{ijcai25}

\newpage
\section{Appendix}
\subsection{Evaluation of LogiDebrief: User Study}
\label{app:user}

To assess the impact of LogiDebrief on emergency call-taking performance, we conducted a structured user study with 27 participants. These participants included individuals with varying levels of experience in emergency call handling: 70.83\% (\textit{n} = 19) were trainees with limited exposure to real-world 9-1-1 call-taking; 25.00\% (\textit{n} = 7) had intermediate experience, having completed training but with limited field application; and 4.17\% (\textit{n} = 1) had prior experience as active 9-1-1 call-takers. Each participant was given a set of emergency call scenarios and underwent debriefing using two different methods: 
(1) traditional human-led quality assurance, and (2) LogiDebrief's automated debriefing system. Evaluations were performed across five key dimensions, rated on a 5-point Likert scale (1: Poor, 5: Excellent). To eliminate bias and ensure a fair comparison, all results from LogiDebrief and human-led debriefing are presented in the same format. The study is distributed in a double-blind manner: participants do not explicitly know which one is from LogiDebreif or human debriefing. The results are summarized in Table~\ref{tab:user_study}.

\begin{table*}[ht]
    \centering
    \caption{Comparison of Traditional Quality Assurance vs. LogiDebrief in User Study (Mean Scores)}
    \label{tab:user_study}
    \begin{tabular}{lcc}
        \toprule
        \textbf{Evaluation Metric} & \textbf{Traditional quality assurance} & \textbf{LogiDebrief} \\
        \midrule
        Actionability (clarity of feedback) & 2.79 & \textbf{4.46}\\
        Comprehensiveness (coverage of key steps) & 2.37 & \textbf{4.33}\\
        Helpfulness (usefulness for skill improvement) & 2.42 & \textbf{4.42}\\
        Self-explanatory nature (ease of understanding) & 2.88 & \textbf{4.38}\\
        Overall preference (percentage favoring method) & 7.41\% & \textbf{92.59\%} \\
        \bottomrule
    \end{tabular}
\end{table*}

The results indicate a strong preference for LogiDebrief, with a substantial improvement in all evaluated dimensions~\citep{milani2024explainable}. In particular: \textbf{Actionability}: Participants found LogiDebrief’s structured feedback significantly more actionable (+59.86\% improvement over human quality assurance). \textbf{Comprehensiveness}: Automated debriefing covered key procedural elements much more effectively (+82.70\% improvement). \textbf{Helpfulness}: Participants perceived LogiDebrief as significantly more beneficial for skill enhancement (+75.21\% improvement). \textbf{Self-explanatory Nature}: LogiDebrief provided clearer, step-by-step feedback (+52.08\% improvement). \textbf{Overall Preference}: 92.59\% of participants preferred LogiDebrief over traditional quality assurance, citing its efficiency and structured guidance.

\textbf{Participant Feedback.} Qualitative feedback reinforced these findings:
\begin{quote}
    \textit{“It walked me through each step instead of just flagging mistakes, so I knew exactly what went wrong and how to fix it.”} 
\end{quote}
\begin{quote}
    \textit{“The automated feedback was fast, so I could complete more simulated calls in the same time and improve much faster.”} 
\end{quote}

These findings suggest that LogiDebrief not only enhances procedural compliance but also improves call-taker learning efficiency by providing real-time, structured feedback. The system's automated nature enables immediate corrections, reducing cognitive overload associated with delayed manual reviews.

\subsection{Scalability and Complexity of LogiDebrief}
\label{app:complex}
We discuss and analyze LogiDebrief's scalability and complexity in this section.

\subsubsection{Real-world Scalability}
LogiDebrief remains \textbf{scalable} by leveraging LLMs to dynamically interpret new emergency scenarios without requiring modifications to its underlying STL framework. Although beyond the scope of this work, call-taking guidelines can be systematically translated into logical specifications using existing tools~\citep{chen2022cityspec, cosler2023nl2spec}, seamlessly integrating procedural updates. Furthermore, the methodology generalizes beyond emergency response training, extending to structured compliance audits in domains such as medical triage and legal training.

\subsubsection{Time Complexity Analysis}
LogiDebrief maintains a worst-case time \textbf{complexity} of \( \mathcal{O}(N) \), where \( N \) denotes the number of conversational turns in an emergency call, previously referred to as \( t \) in \( \omega \). This complexity arises from three main processing stages: contextual understanding, runtime verification, and aggregation. We introduce additional notations to reflect the influence of different factors on complexity and scalability. Contextual understanding involves extracting key details such as responder type, call category, and critical conditions from the call transcript. Given that emergency protocols define a finite set of responder types (\( R \)) and call categories (\( C \)), the complexity remains:
\begin{equation}
\mathcal{O}(N \cdot |R| + N \cdot |C|)
\end{equation}
Since \( |R| \) and \( |C| \) are bounded constants, this phase simplifies to \( \mathcal{O}(N) \). The worst case occurs when all responder types and categories need to be considered, but the system optimizes this by filtering irrelevant checks early.
Runtime verification applies procedural compliance checks over the conversational transcript. Let \( K \) be the number of compliance checks per call, \( P \) be the number of protocol documents referenced, and \( M \) be the number of retrieval operations needed from external knowledge.
Each compliance check runs over a relevant subset of turns (\( N' \leq N \)), leading to a worst-case complexity:
\begin{equation}
\mathcal{O}(K \cdot N' + P \cdot M)
\end{equation}
Since \( K \) is bounded by emergency response protocols and \( M \) is constrained by RAG efficiency, the worst-case remains \( \mathcal{O}(N) \).
Aggregation involves classifying compliance into discrete categories (Yes, No, Refused, NA). Given that each check results in a predefined category, aggregation requires only a constant number of operations per check:
\begin{equation}
\mathcal{O}(K)
\end{equation}
Since \( K \) is bounded, aggregation runs in \( \mathcal{O}(1) \).
Overall, summing the complexity from all three stages, the worst-case complexity is:
\begin{equation}
\mathcal{O}(N) + \mathcal{O}(N) + \mathcal{O}(1) = \mathcal{O}(N)
\end{equation}

\subsubsection{Runtime Efficiency in Practice}

We select 200 calls from our dataset, spanning durations from 2 to 20 minutes. While practical runtime efficiency may vary due to factors such as call complexity, machine configurations, coding implementations, and internet connectivity, we report the following findings based on the time spent and standard deviation under multiple LogiDebrief runtime:  
\begin{itemize}
    \item Calls (2–10 min): \quad 5.81$\pm$1.06 sec per call minute  
    \item Calls (10–15 min): \quad 5.61$\pm$1.20 sec per call minute  
    \item Calls (15–20 min): \quad 5.93$\pm$1.67 sec per call minute  
\end{itemize}
\noindent Overall, LogiDebrief maintains a consistent runtime efficiency across different call lengths, with only slight variations under practical runtimes with real-world data.

\subsubsection{Long-term Sustainability}
One key challenge in automated debriefing is ensuring alignment between AI-driven assessments and human quality assurance (quality assurance) evaluations. In real-world deployment, we observed cases where LogiDebrief flagged procedural deviations overlooked by human reviewers, leading to enhanced protocol compliance. However, to maintain trust and adaptability, future iterations will incorporate human-in-the-loop (HITL) mechanisms, allowing quality assurance officers to refine STL-based rules dynamically. Further user studies are planned to evaluate long-term behavioral impacts on call-takers.

\subsection{Algorithmic Description}
\label{app:algo}

In this section, we introduce the algorithmic description in Alg. \ref{alg:logi_debrief} of LogiDebrief understanding the call context.

\subsection{Extended Related Work}
\label{app:related}

\subsubsection{Large Language Models (LLMs) in Procedural Verification} Large Language Models (LLMs) have demonstrated strong capabilities in natural language understanding and reasoning. However, their application in procedural verification remains a significant challenge. While Chain-of-Thought (CoT) prompting enhances reasoning by breaking down multi-step problems~\citep{wei2022chain}, it does not guarantee procedural adherence, often leading to logical inconsistencies and hallucinations~\citep{miao2023selfcheck}. Retrieval-Augmented Generation (RAG)~\citep{shi2023replug} improves factual accuracy by incorporating external knowledge. However, retrieval mechanisms are prone to errors, often failing to retrieve the most relevant procedural guidelines or synthesizing information across multiple sources~\citep{shuster2022language}. This limitation makes RAG unreliable for high-stakes verification, such as emergency call debriefing, where missing a single protocol step can compromise evaluation. Self-verification techniques, where LLMs assess their own outputs for consistency~\citep{zelikman2022star}, improve coherence but do not enforce procedural correctness. Additionally, as context lengths increase, LLMs struggle to integrate long-range dependencies effectively, leading to incomplete procedural evaluations~\citep{liu2024lost}. These challenges highlight the need for structured, rule-based verification frameworks, such as Signal Temporal Logic (STL), to enhance procedural reliability.

\subsubsection{Reasoning Frameworks} Recent reasoning-focused LLM frameworks aim to improve structured decision-making and logical consistency. OpenAI-o1~\citep{openai2024o1} is optimized for step-by-step problem solving, incorporating reinforcement learning techniques to improve logical inference. While OpenAI-o1 excels in symbolic reasoning and stepwise verification, it still encounters difficulties in real-time procedural compliance checks, particularly in environments where external document retrieval is necessary. DeepSeek-r1~\citep{deepseek2025r1} is specifically trained to enhance logical coherence through reinforcement learning from reasoning feedback (RLRF). Unlike traditional LLMs, DeepSeek-r1 incentivizes structured reasoning by refining its thought process iteratively. Despite outperforming general-purpose LLMs in logical inference tasks, it still lacks formal compliance enforcement mechanisms, making it less reliable for structured regulatory evaluations such as 9-1-1 call debriefing. Gemini-Flash-Thinking, an experimental model from Google DeepMind~\citep{Google2024Gemini}, is designed for multi-modal reasoning with external memory integration. This approach enhances its ability to reference and retrieve procedural documents dynamically. However, Gemini-Flash-Thinking exhibits contextual drift, where retrieved documents influence responses inconsistently, leading to procedural misinterpretation in complex multi-step evaluations. While OpenAI-o1 and DeepSeek-r1 provide advancements in structured reasoning, none fully address procedural verification challenges in high-stakes environments. These frameworks lack rule-based validation mechanisms and are susceptible to retrieval errors or prompt degradation in long-form evaluations. LogiDebrief bridges this gap by integrating LLMs with STL-based runtime monitoring, ensuring logical consistency and compliance with procedural guidelines. Unlike ICL+RAG approaches that rely on implicit reasoning, LogiDebrief formally encodes procedural rules, enabling deterministic verification of call-taking adherence. Proven by experimental results, this hybrid approach outperforms standalone reasoning models by combining linguistic flexibility with structured logic enforcement.

\subsection{Prompting Templates}
\label{app:prompt}
\subsubsection{Prompt Templates for Functions}
To integrate LLMs into STL functions, we design structured prompting templates to ensure consistency, efficiency, and interpretability. Each template explicitly defines the input structure, expected output format, and its role in procedural verification. To mitigate LLM-induced \textbf{hallucinations}, LogiDebrief employs a confidence-aware fallback mechanism. If an LLM-generated check yields an in-context confidence score below 70\% (assessed within prompts but omitted from subsequent templates for readability), the system flags the result as `low confidence' and escalates it for human review. Additionally, all LLM-generated outputs are systematically structured and refined to enhance accuracy and alignment with procedural requirements.

\begin{algorithm}  % Use [H] to prevent floating to next column
\footnotesize  % Reduce font size to better fit double-column format
\caption{Contextual Understanding in LogiDebrief}
\label{alg:logi_debrief}

\begin{algorithmic}[1]

\STATE \textbf{Input:} 911 Call $\omega$
\STATE \textbf{Output:} Quality assurance form $\Psi$

\STATE $\hat{\mathrm{R}} \gets \emptyset$
\FOR{$\mathrm{r} \in \{\mathrm{fire, police, medical}\}$}
    \IF{$\texttt{SCENE}(\omega, r) = \top$}
        \STATE $\hat{\mathrm{R}} \gets \hat{\mathrm{R}} \cup \{r\}$
    \ENDIF
\ENDFOR

\STATE $\hat{\mathrm{T}} \gets \emptyset$
\FOR{$\mathrm{t} \in \{\mathrm{Call \ Types}\mid{\hat{\mathrm{R}}}\}$}
    \IF{$\texttt{TYPE}(\omega, t) = \top$}
        \STATE $\hat{\mathrm{T}} \gets \hat{\mathrm{T}} \cup \{t\}$ 
    \ENDIF
\ENDFOR

\STATE $\hat{\mathrm{C}} \gets \emptyset$
\FOR{$\mathrm{c} \in \{\mathrm{Criticals}_{\times 6}\}$}
    \IF{$\texttt{CRITICAL}(\omega, c) = \top$}
        \STATE $\hat{C} \gets \hat{C} \cup \{c\}$ 
    \ENDIF
\ENDFOR

\STATE $\Psi \gets \Gamma(\mathrm{\hat{R}})$
\FOR{$\varphi \in \Psi$}
    \STATE $\varphi \gets \varphi \oplus \Delta(\hat{\mathrm{T}}, \hat{\mathrm{C}})$
\ENDFOR

\FOR{$\varphi \in \Psi$}
    \FOR{$r \in \varphi$}
        \STATE $\mathbb{I}(\mathcal{P} \mid{r}) \gets \top$
        \FOR{$p \in \{\mathcal{P} \mid{r}\}$}
            \STATE $\mathbb{I}(\mathcal{P}\mid{r}) \gets \mathbb{I}(\mathcal{P} \mid{r}) \bigwedge \texttt{SCAN}(\omega, p)$
        \ENDFOR
        \IF{$\mathbb{I}(\mathcal{P}\mid{r}) = \bot$}
            \STATE $\Psi \gets \Psi - \{r\}$
        \ENDIF
    \ENDFOR
\ENDFOR

\end{algorithmic}
\end{algorithm}

\paragraph{\texttt{SCENE} Function}

\noindent Prompt Template:
\begin{tcolorbox}
\textbf{[System]:} Identify whether the emergency scenario requires a Fire, Police, or Medical responder based on the following conversation transcript. 

\texttt{\{\textit{Few-shot examples here}\}}

\textbf{[Call Transcript]:}  
\texttt{\{CALL\_TRANSCRIPT\}}

\textbf{[Task]:} Based on the conversation, determine the required responders. Does it require Fire/Police/Medical? Return only Yes or No.
\end{tcolorbox}

\noindent Expected Output:
\begin{verbatim}
Yes
\end{verbatim}
or  
\begin{verbatim}
No
\end{verbatim}

\noindent STL Function:
\begin{equation}
\texttt{SCENE}(\omega, \text{responders}) \coloneqq 
\Diamond_{[0,T]} \big(\omega(t) \models \text{responders}\big)
\end{equation}

\paragraph{\texttt{TYPE} Function}

\noindent Prompt Template:
\begin{tcolorbox}
\textbf{[System]:} Classify the emergency type in the following conversation transcript. 

\texttt{\{\textit{Few-shot examples here}\}}

\textbf{[Call Transcript]:}  
\texttt{\{CALL\_TRANSCRIPT\}}

\textbf{[Task]:} Identify if the call belongs to this \texttt{\{incident type here\}}? Return only Yes or No.
\end{tcolorbox}

\noindent Expected Output:
\begin{verbatim}
Yes
\end{verbatim}
or  
\begin{verbatim}
No
\end{verbatim}

\noindent STL Function:
\begin{equation}
\texttt{TYPE}(\omega, \text{type}) \coloneqq 
\Diamond_{[0,T]} \big(\omega(t) \models \text{type}\big)
\end{equation}

\paragraph{\texttt{CRITICAL} Function}

\noindent Prompt Template:
\begin{tcolorbox}
\textbf{[System]:} Identify whether the emergency involves a critical life-threatening situation. 

\texttt{\{\textit{Few-shot examples here}\}}

\textbf{[Call Transcript]:}  
\texttt{\{CALL\_TRANSCRIPT\}}

\textbf{[Task]:} Determine if \texttt{\{critical situation\}} is present.

Return only Yes and No.
\end{tcolorbox}

\noindent Expected Output:
\begin{verbatim}
Yes
\end{verbatim}
or  
\begin{verbatim}
No
\end{verbatim}

\noindent STL Function:
\begin{equation}
\texttt{CRITICAL}(\omega, \text{flag}) \coloneqq 
\Diamond_{[0,T]} \big(\omega(t) \models \text{flag} \big)
\end{equation}

\paragraph{\texttt{DETECT} Function}

\noindent Prompt Template:
\begin{tcolorbox}
\textbf{[System]:} Verify whether the call-taker performed a required action. 

\texttt{\{\textit{Few-shot examples here}\}}

\textbf{[Call Transcript]:}  
\texttt{\{CALL\_TRANSCRIPT\}}

\textbf{[Task]:} Determine whether the call-taker performed the following action:  
"\texttt{\{ACTION\_TO\_VERIFY\}}"

Return ``Yes'' if performed, otherwise return ``No."
\end{tcolorbox}

\noindent Expected Output:
\begin{verbatim}
Yes
\end{verbatim}
or  
\begin{verbatim}
No
\end{verbatim}

\noindent STL Function:
\begin{equation}
\texttt{DETECT}(\omega, \text{action}) \coloneqq 
\Diamond_{[0,T]} \big(\omega(t) \models \text{action} \big)
\end{equation}

\paragraph{\texttt{SCAN} Function}

\noindent Prompt Template:
\begin{tcolorbox}
\textbf{[System]:} Verify whether a specific precondition is met in the following conversation transcript. 

\texttt{\{\textit{Few-shot examples here}\}}

\textbf{[Call Transcript]:}  
\texttt{\{CALL\_TRANSCRIPT\}}

\textbf{[Task]:} Determine whether the following precondition is satisfied:  
"\texttt{\{PRECONDITION\}}"

Return ``Yes'' if the precondition is met at any point in the conversation, otherwise return ``No.''
\end{tcolorbox}

\noindent Expected Output:
\begin{verbatim}
Yes
\end{verbatim}
or  
\begin{verbatim}
No
\end{verbatim}

\noindent STL Function:
\begin{equation}
\texttt{SCAN}(\omega, \text{precondition}) \coloneqq 
\Diamond_{[0,T]} \big(\omega(t) \models \text{precondition} \big)
\end{equation}

\paragraph{\texttt{answer} Function}

\noindent Prompt Template:
\begin{tcolorbox}
\textbf{[System]:} Extract specific details from the call transcript. 

\texttt{\{\textit{Few-shot examples here}\}}

\textbf{[Call Transcript]:}  
\texttt{\{CALL\_TRANSCRIPT\}}

\textbf{[Task]:} Extract the following information:  
"\texttt{\{QUESTION\}}"

Return the exact response. If unavailable, return ``N/A."
\end{tcolorbox}

\noindent Expected Output:
\begin{verbatim}
"123 Main Street, Apt 4B"
\end{verbatim}
or  
\begin{verbatim}
"N/A"
\end{verbatim}

\noindent STL Function:
\begin{equation}
\texttt{ANSWER}(\omega, \text{query}) \rightarrow a
\end{equation}

\subsubsection{Prompt Templates for Evaluation}

\noindent Prompt Template:
\begin{tcolorbox}[width=\columnwidth]
\textbf{[System]:} You are an expert in 9-1-1 call debriefing. Assess compliance using the following procedural manuals and structured reasoning.

\textbf{[Retrieved Manuals]:}  \\
\texttt{\{RETRIEVED\_CALL\_TAKING\_MANUAL\}}

\textbf{[Few-Shot Examples]:}  \\
\texttt{\{\textit{Examples of structured debriefing and Chain-of-Thought reasoning}\}}

\textbf{[Call Transcript]:}  
\texttt{\{CALL\_TRANSCRIPT\}}

\textbf{[Task]:}  
Compare the transcript against procedural guidelines for the requirement:  
``\texttt{\{REQUIREMENT\}}''

\textbf{Reasoning Steps:}  \\
(1) Identify relevant steps from manuals.  \\
(2) Compare with the call-taker’s actions.  \\
(3) Determine compliance and explain reasoning.  \\ 
(4) If non-compliant, suggest corrective actions.

\textbf{Expected Output:  }\\
Compliance: (Yes/No/NA/Caller Refused) \\
Rationale: \texttt{Explanation}
\end{tcolorbox}

\noindent Example Output:
\begin{tcolorbox}[width=\columnwidth]
\textbf{Compliance: Yes}  \\
\textbf{Rationale:} The call-taker verified the address by confirming it twice,  
cross-checking with geographic data, and recording it before disconnection.
\end{tcolorbox}

or  

\begin{tcolorbox}[width=\columnwidth]
\textbf{Compliance: No}  \\
\textbf{Rationale:} The call-taker failed to reconfirm the address before disconnection.  
To improve, they should always perform a final confirmation check.
\end{tcolorbox}

\subsection{Ethical Considerations}

This work adheres to strict ethical guidelines to ensure safety, privacy, and responsible AI deployment. Below, we outline key ethical considerations:  

\noindent \textbf{1. IRB Approval:}  
This research has been approved by the Institutional Review Board (IRB) of the authors' affiliated research institutions, ensuring compliance with ethical standards for human subject research.  

\noindent \textbf{2. Collaborative Development with Government Agencies:}  
LogiDebrief is a result of collaboration between researchers and governmental agencies, following an iterative design and development loop. Ethical concerns, including safety, security, and privacy, have been proactively addressed throughout the development process.  

\noindent \textbf{3. Non-Intrusive Deployment:}  
During real-world deployment, LogiDebrief does not interact with callers or call-takers during active calls. It operates post-call, delivering just-in-time debriefing results and instructions without interrupting live emergency response operations.  

\noindent \textbf{4. Training-Only Use Case:}  
LogiDebrief is strictly used for training purposes and does not have access to real-time call-taking or dispatching systems. It is designed to assist in quality assurance and skill development rather than influence active emergency response decisions.  

\noindent \textbf{5. User Control and Monitoring:}  
LogiDebrief operates under continuous monitoring and can be terminated at any time if users feel uncomfortable or unsafe interacting with AI. This ensures transparency, user autonomy, and responsible AI deployment in emergency communication training.  

By adhering to these ethical guidelines, LogiDebrief ensures responsible AI integration while supporting the development of well-trained emergency response personnel.

\end{document}